%% file: acl_latex.tex
\documentclass[11pt]{article}

\usepackage[preprint]{acl}
\usepackage{times}
\usepackage{latexsym}
\usepackage{amsmath}
\usepackage{tikz}
\usetikzlibrary{positioning,calc,arrows.meta,shapes.geometric}
\usepackage[T1]{fontenc}

\usepackage[utf8]{inputenc}

\usepackage{microtype}

\usepackage{inconsolata}

\usepackage{graphicx}
\usepackage{microtype}
\usepackage{hyperref}
\usepackage{url}
\usepackage{booktabs}
\usepackage{graphicx}
\usepackage{amsmath}
\usepackage{float}

\usepackage{lineno}
\usepackage[textsize=tiny]{todonotes}
\usepackage{enumitem}
\usepackage{xcolor}
\usepackage{colortbl}
\usepackage{multirow}
\definecolor{LightGray}{gray}{0.5}
\definecolor{kh}{HTML}{168aff}

\definecolor{nk}{HTML}{9B59B6}

\definecolor{darkblue}{rgb}{0, 0, 0.5}
\hypersetup{colorlinks=true, citecolor=darkblue, linkcolor=darkblue, urlcolor=darkblue}
\newcommand{\llamamodel}{Llama-3.1-8B-Instruct}
\newcommand{\qwenseven}{Qwen2.5-7B-Instruct}
\newcommand{\qwenfourteen}{Qwen2.5-14B-Instruct}

\input{math_commands}

\title{Language Steering for Multilingual In-Context Learning}

\author{Neeraja Kirtane\quad Kuan-Hao Huang \vspace{0.1em}\\
Texas A\&M University\vspace{0.1em}\\
\texttt{\{neeraja,khhuang\}@tamu.edu}}

\begin{document}
\maketitle
\begin{abstract}
If multilingual language model representations contain separable language-associated structure, then shifting between languages may require only simple activation-space interventions.
To test this hypothesis, we study multilingual in-context learning (ICL) as a controlled setting: few-shot demonstrations are provided in English while the test query is in a target language, creating a clear cross-lingual transfer challenge. We propose language vectors, which are computed as mean activation differences between parallel source and target language examples at a particular layer and applied as inference-time offsets to the model's hidden states. Across three tasks spanning 19 languages and three models, our method improves over English few-shot prompting without any parameter updates. Beyond performance, the vectors encode interpretable linguistic structure: closely related languages form tight clusters, and vectors transfer effectively across tasks and languages, with closely related languages' vectors recovering more of the gain than distant ones. Together, these findings suggest that language occupies structured directions in multilingual activation space, and that inference-time interventions on English demonstrations can narrow the cross-lingual ICL gap.

\end{abstract}

\section{Introduction}

\input{introduction}
\input{methodology}

\section{Experiments}
\input{results_new}
\section{What Do Language Vectors Encode?}
\label{ablation}
\input{ablation}

\section{Related Work}
\input{related_work}

\section{Conclusion}
\input{conclusion}

\bibliography{custom}

\clearpage

\appendix
\input{appendix}

\end{document}

%% file: math_commands.tex
\usepackage{amsmath,amsfonts,bm}

\def\eqref#1{equation~\ref{#1}}

\def\1{\bm{1}}

\DeclareMathAlphabet{\mathsfit}{\encodingdefault}{\sfdefault}{m}{sl}
\SetMathAlphabet{\mathsfit}{bold}{\encodingdefault}{\sfdefault}{bx}{n}



%% file: introduction.tex
Large language models have demonstrated remarkable multilingual capabilities, yet a persistent performance gap between English and low-resource languages limits equitable access across downstream tasks \cite{zhao2024large, qin2024multilingual}. A key question is whether this gap partly reflects how multilingual information is represented and accessed during inference. If language-associated variation can be identified in activation space, interventions along these directions may help improve cross-lingual transfer.

\begin{figure*}
    \centering
    \includegraphics[width=\textwidth]{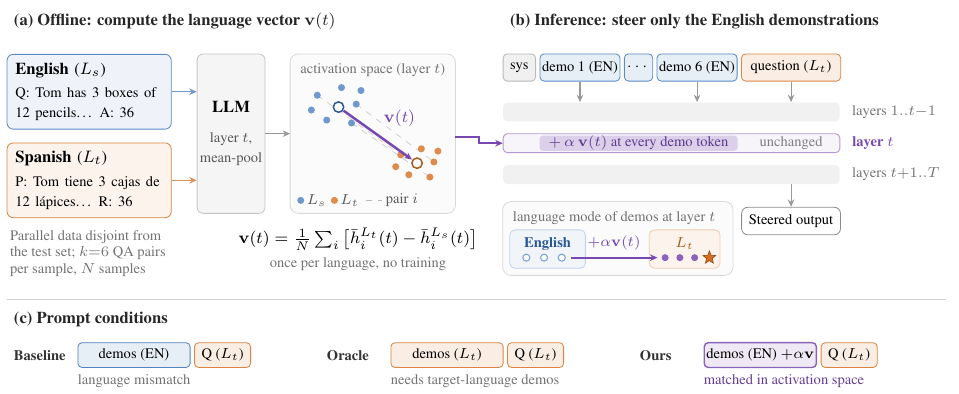}
    \caption{Overview of language steering for multilingual in-context learning. \textbf{(a) Offline vector computation:} parallel English ($L_s$) and target-language ($L_t$)
question-answer sequences are passed through the LLM, and their hidden states at layer $t$ are
mean-pooled. The language vector $\mathbf{v}(t)$ is the mean of the paired differences across
$N$ samples. It is computed once per language, with no
parameter updates.
\textbf{(b) Inference-time steering:} given English demonstrations and a target-language
question, we add $\alpha\ \mathbf{v}(t)$ to the hidden states of the demonstration tokens at
layer $t$, leaving the question tokens unchanged (more configs detailed in Section \ref{sec:inference_steer}).
\textbf{(c) Prompt conditions:} the baseline pairs English demonstrations with a
target-language question, the oracle uses target-language demonstrations, and our method
matches the two languages in activation space.}
    \label{fig:placeholder}
\end{figure*}

Recent work in mechanistic interpretability offers a reason to believe this 
redirection is possible. Transformer-based models have been shown to encode 
distinct patterns for different types of information within their internal 
representations \cite{rai2024practical, li2024improving, elhage2022superposition}, 
suggesting that concepts are not uniformly distributed but occupy structured, 
separable directions in activation space. This raises a natural question: is 
language identity encoded the same way? If so, a simple activation offset should suffice to steer the model toward target-language processing. Indeed, mean 
activation differences between contrastive inputs have been shown to encode 
interpretable directions in representation space \cite{marks2023geometry, 
rimsky2024steering, park2023linear}, and such directions have been applied to address language 
confusion in multilingual models \cite{yunfan2025mitigating, 
sterz2025recover}. However, these approaches primarily focus
on output-language fidelity. Whether
language-associated activation directions
can also improve downstream cross-lingual
task performance remains less understood.

To investigate this, we need a setting with clear controllability over the 
language of inputs and outputs. In-context learning (ICL) \citep{brown2020language} naturally provides 
this. When demonstrations are provided in a source language (English) where 
high-quality data exists but the test query is in a low-resource target 
language, models must internally recognise the language switch and transfer 
their task understanding across languages,   a computationally and 
representationally demanding process that often fails. This cross-lingual ICL gap provides a natural testbed for our hypothesis.  

To test this hypothesis, we apply \textit{language vectors}, mean activation differences between hidden states of parallel source and target language examples, added as inference-time offsets, to the multilingual ICL setting.
As shown in Figure~\ref{fig:placeholder}, our method operates 
in a few-shot ICL setting, where we: (1) collect paired examples in the source 
and target language, (2) compute the activation difference to extract a 
language-specific steering vector, and (3) apply this vector during inference 
to shift the model's internal representations toward target-language processing behavior.
We use language vectors to investigate
whether language-associated activation
differences exhibit interpretable structure
and can be leveraged to improve cross-lingual
ICL performance. Vectors computed from a different dataset or task, or from general-purpose parallel sentences, improve performance (Section~\ref{sec:cross_task_transfer}). 

We evaluate across three datasets spanning mathematical reasoning and natural language inference, 19 languages, and three models. Our results support the hypothesis and show that language vectors are reusable rather than tied to the data they were computed from. A single vector per language, computed once offline, improves over the cross-lingual baseline in every model--dataset setting, and the vectors generalize along three axes. \emph{Across tasks}, vectors computed on one task transfer to others. \emph{Across datasets}, vectors computed on a different math dataset improve MGSM nearly as much as in-benchmark vectors (65.9 vs.\ 67.9), and vectors computed from general-purpose parallel sentences also improve over the baseline on MGSM and MSVAMP. \emph{Across languages}, a closely related language's vector recovers more of the gain than distant or random vectors. Finally, the vectors encode interpretable linguistic structure: they cluster by family, typology, and script, with closely related pairs merging together.

The contributions of our work are: (1) We apply mean-difference activation steering \citep{marks2023geometry, park2023linear, liu2023context} to cross-lingual ICL as a probe of whether language identity occupies exploitable directions in multilingual activation space.
(2) We show that steering English demonstrations with these vectors improves cross-lingual ICL across three datasets, 19 languages, and three models. (3) We show that the vectors are reusable: a single vector per language generalizes across tasks, datasets, and closely related languages. (4) We show that the vectors encode interpretable linguistic structure, clustering by family, typology, and script.

%% file: methodology.tex
\section{Language Vectors: Method}
We consider multilingual in-context learning as our setup following \citet{tu2025blessing}, where each instance consists of few-shot source (English) demonstrations and a test question in the target language, and aim to improve over this cross-lingual baseline toward the performance achievable with target-language demonstrations.

\subsection{Problem Formulation and Notations}




Let $L_s$ denote the source language and $L_t$ denote the target language. In the multilingual in-context learning setup, the model receives a prompt consisting of $k$ few-shot demonstration examples $\mathcal{F} = \{(q_i^{L_s}, c_i^{L_s}, a_i^{L_s})\}_{i=1}^k$ followed by a test question $q_{\text{test}}^{L_t}$. Each demonstration includes a question $q_i^{L_s}$, chain-of-thought reasoning $c_i^{L_s}$ (if any), and answer $a_i^{L_s}$, all in the source language $L_s$. 
The model must use task information conveyed by the source-language demonstrations to predict the answer to the target-language query $q_{\text{test}}^{L_t}$.
Given a language model $\mathcal{M}$ with $T$ layers, we denote the hidden state at layer $t \in \{1, \ldots, T\}$ and token position $j$ for input text $x$ as $\mathbf{h}_j(t) \in \mathbb{R}^d$, where $d$ is the model's hidden state size.

\subsection{Computing Language Directions}
\label{2.2}

Our goal is to compute a language-specific steering vector $\mathbf{v}(t) \in \mathbb{R}^d$ that, when added to hidden states during inference, improves the model's performance on target language inputs $q^{L_t}$ beyond the baseline. We fix the source language as English and compute steering vectors from activation patterns across parallel source-target examples.



\paragraph{Few-Shot Format for Language Vector Computation.}
To capture richer contextual 
patterns, we construct $N = |\mathcal{D}_{\text{compute}}|$ samples, each 
consisting of $k$ question-answer pairs concatenated together. Specifically, 
for each example $i \in \mathcal{D}_{\text{compute}}$, we construct a sample 
by placing example $i$ in the first slot and filling the remaining $k-1$ slots 
by sampling uniformly with replacement from $\mathcal{D}_{\text{compute}}$. 
This ensures each example appears at least once across all constructed samples. 
For each sample $i$ in our compute set $\mathcal{D}_{\text{compute}}$, we 
format parallel texts as:
{
\begin{equation}
x^{L_s}_i = \bigoplus_{m=1}^{k} [q^{L_s}_{i,m}, a^{L_s}_{i,m}], \quad
x^{L_t}_i = \bigoplus_{m=1}^{k} [q^{L_t}_{i,m}, a^{L_t}_{i,m}]
\end{equation}
}

\noindent where $q_{i,m}^{L_s}$ and $q_{i,m}^{L_t}$ are parallel questions in source and target languages respectively, $a_{i,m}^{L_s}$ and $a_{i,m}^{L_t}$ are the answers. For datasets with chain-of-thought reasoning, both sequences contain the
same English reasoning $c_{i,m}^{L_s}$.
$\oplus$ denotes concatenation, and $k$ is the number of question-answer pairs per sample.


\paragraph{Activation Extraction and Vector Computation.}
We extract hidden states at layer $t$ for both source and target formatted texts. Recall that each sample $i$ is a concatenated sequence of $k$ question-answer pairs. We pass each such sequence through model M and mean-pool the hidden states across all token positions to obtain a single vector per sample.

\begin{equation}
    \mathbf{h}_i^{L_s}(t) = \frac{1}{|x_i^{L_s}|} \sum_{j=1}^{|x_i^{L_s}|} \mathbf{h}_{i,j}^{L_s}(t)
\end{equation}
\begin{equation}
    \mathbf{h}_i^{L_t}(t) = \frac{1}{|x_i^{L_t}|} \sum_{j=1}^{|x_i^{L_t}|} \mathbf{h}_{i,j}^{L_t}(t)
\end{equation}

\noindent where $\mathbf{h}_{i,j}^{L_s}(t)$ denotes the hidden state at token position $j$ of sample $i$ in the source language at layer $t$, and $|x_i^{L_s}|$ is the total sequence length. We compare mean pooling against last-token extraction in Appendix~\ref{appendix:pooling}, finding mean pooling consistently superior.
\begin{table*}[t]
\centering
\setlength{\tabcolsep}{6pt}
\resizebox{0.92\textwidth}{!}{
\begin{tabular}{lcccccccccccc}
\toprule
\multirow{2}{*}{Language} & \multicolumn{4}{c}{MGSM} & \multicolumn{4}{c}{XNLI} & \multicolumn{4}{c}{MSVAMP} \\
\cmidrule(lr){2-5}\cmidrule(lr){6-9}\cmidrule(lr){10-13}
& B & MFS & Ours & \cellcolor{LightGray!30}OR 
& B & MFS & Ours & \cellcolor{LightGray!30}OR 
& B & MFS & Ours & \cellcolor{LightGray!30}OR \\
\midrule
Arabic      & \textcolor{gray}{--} & \textcolor{gray}{--} & \textcolor{gray}{--} & \cellcolor{LightGray!30}\textcolor{gray}{--} & 62.57 & 60.78 & \bf{62.87} & \cellcolor{LightGray!30}63.77 & \textcolor{gray}{--} & \textcolor{gray}{--} & \textcolor{gray}{--} & \cellcolor{LightGray!30}\textcolor{gray}{--} \\
Basque      & 32.14 & 35.71 & \bf{36.90} & \cellcolor{LightGray!30}52.38 & \textcolor{gray}{--} & \textcolor{gray}{--} & \textcolor{gray}{--} & \cellcolor{LightGray!30}\textcolor{gray}{--} & \textcolor{gray}{--} & \textcolor{gray}{--} & \textcolor{gray}{--} & \cellcolor{LightGray!30}\textcolor{gray}{--} \\
Bengali     & 57.14 & 55.95 & \bf{61.90} & \cellcolor{LightGray!30}58.33 & \textcolor{gray}{--} & \textcolor{gray}{--} & \textcolor{gray}{--} & \cellcolor{LightGray!30}\textcolor{gray}{--} & 57.49 & \bf{62.87} & 59.58 & \cellcolor{LightGray!30}61.38 \\
Bulgarian   & \textcolor{gray}{--} & \textcolor{gray}{--} & \textcolor{gray}{--} & \cellcolor{LightGray!30}\textcolor{gray}{--} & 56.29 & \bf{61.98} & 61.68 & \cellcolor{LightGray!30}66.17 & \textcolor{gray}{--} & \textcolor{gray}{--} & \textcolor{gray}{--} & \cellcolor{LightGray!30}\textcolor{gray}{--} \\
Catalan     & 64.29 & 64.29 & \bf{69.05} & \cellcolor{LightGray!30}76.19 & \textcolor{gray}{--} & \textcolor{gray}{--} & \textcolor{gray}{--} & \cellcolor{LightGray!30}\textcolor{gray}{--} & \textcolor{gray}{--} & \textcolor{gray}{--} & \textcolor{gray}{--} & \cellcolor{LightGray!30}\textcolor{gray}{--} \\
Chinese     & 67.86 & 67.86 & \bf{71.43} & \cellcolor{LightGray!30}72.62 & 59.88 & \bf{61.38} & 59.28 & \cellcolor{LightGray!30}61.98 & 69.76 & \bf{73.05} & 71.26 & \cellcolor{LightGray!30}73.35 \\
French      & 61.90 & 64.29 & \bf{70.24} & \cellcolor{LightGray!30}65.48 & 67.37 & 68.26 & \bf{72.75} & 7\cellcolor{LightGray!30}1.26 & 71.56 & 73.05 & \bf{74.55} & \cellcolor{LightGray!30}73.05 \\
Galician    & 64.29 & 69.05 & \bf{73.81} & \cellcolor{LightGray!30}77.38 & \textcolor{gray}{--} & \textcolor{gray}{--} & \textcolor{gray}{--} & \cellcolor{LightGray!30}\textcolor{gray}{--} & \textcolor{gray}{--} & \textcolor{gray}{--} & \textcolor{gray}{--} & \cellcolor{LightGray!30}\textcolor{gray}{--} \\
German      & 66.67 & 66.67 & \bf{75.00} & \cellcolor{LightGray!30}71.43 & 64.07 & 65.27 & \bf{66.47} & \cellcolor{LightGray!30}65.27 & 71.26 & 70.06 & \bf{71.26} & \cellcolor{LightGray!30}76.95 \\
Greek       & \textcolor{gray}{--} & \textcolor{gray}{--} & \textcolor{gray}{--} & \cellcolor{LightGray!30}\textcolor{gray}{--} & 68.26 & 66.17 & \bf{70.06} & \cellcolor{LightGray!30}72.75 & \textcolor{gray}{--} & \textcolor{gray}{--} & \textcolor{gray}{--} & \cellcolor{LightGray!30}\textcolor{gray}{--} \\
Hindi       & \textcolor{gray}{--} & \textcolor{gray}{--} & \textcolor{gray}{--} & \cellcolor{LightGray!30}\textcolor{gray}{--} & 61.38 & 57.78 & \bf{64.97} & \cellcolor{LightGray!30}61.98 & \textcolor{gray}{--} & \textcolor{gray}{--} & \textcolor{gray}{--} & \cellcolor{LightGray!30}\textcolor{gray}{--} \\
Japanese    & 55.95 & \bf{61.90} & 55.95 & \cellcolor{LightGray!30}63.10 & \textcolor{gray}{--} & \textcolor{gray}{--} & \textcolor{gray}{--} & \cellcolor{LightGray!30}\textcolor{gray}{--} & 63.17 & \bf{68.26} & 67.96 & \cellcolor{LightGray!30}70.06 \\
Russian     & 71.43 & 71.43 & \bf{72.62} & \cellcolor{LightGray!30}76.19 & 58.98 & \bf{64.37} & 63.77 & \cellcolor{LightGray!30}60.78 & 68.86 & \bf{72.46} & 72.16 & \cellcolor{LightGray!30}71.56 \\
Spanish     & \bf{77.38} & 70.24 & {76.19} & \cellcolor{LightGray!30}78.57 & 66.77 & 66.17 & \bf{70.66} & \cellcolor{LightGray!30}67.37 & 74.55 & 73.95 & \bf{75.75} & \cellcolor{LightGray!30}74.55 \\
Swahili     & 55.95 & 63.10 & \bf{65.48} & \cellcolor{LightGray!30}66.67 & 52.40 & 51.20 & \bf{55.99} & \cellcolor{LightGray!30}56.89 & 56.29 & 59.58 & \bf{60.78} & \cellcolor{LightGray!30}62.87 \\
Thai        & 57.14 & \bf{67.86} & 61.90 & \cellcolor{LightGray!30}59.52 & 59.88 & \bf{64.37} & 60.78 & \cellcolor{LightGray!30}70.66 & 59.58 & \bf{64.67} & 64.07 & \cellcolor{LightGray!30}66.77 \\
Turkish     & \textcolor{gray}{--} & \textcolor{gray}{--} & \textcolor{gray}{--} & \cellcolor{LightGray!30}\textcolor{gray}{--} & 62.28 & 57.19 & \bf{65.87} & \cellcolor{LightGray!30}65.57 & \textcolor{gray}{--} & \textcolor{gray}{--} & \textcolor{gray}{--} & \cellcolor{LightGray!30}\textcolor{gray}{--} \\
Urdu        & \textcolor{gray}{--} & \textcolor{gray}{--} & \textcolor{gray}{--} & \cellcolor{LightGray!30}\textcolor{gray}{--} & 55.09 & 55.39 & \bf{56.29} & \cellcolor{LightGray!30}57.19 & \textcolor{gray}{--} & \textcolor{gray}{--} & \textcolor{gray}{--} & \cellcolor{LightGray!30}\textcolor{gray}{--} \\
Vietnamese  & \textcolor{gray}{--} & \textcolor{gray}{--} & \textcolor{gray}{--} & \cellcolor{LightGray!30}\textcolor{gray}{--} & 64.07 & 63.17 & \bf{68.86} & \cellcolor{LightGray!30}68.86 & \textcolor{gray}{--} & \textcolor{gray}{--} & \textcolor{gray}{--} & \cellcolor{LightGray!30}\textcolor{gray}{--} \\
\midrule
Average & 61.01 & 63.19 & \bf{65.87} & \cellcolor{LightGray!30}68.16 & 61.38 & 61.68 & \bf{64.31} & \cellcolor{LightGray!30}65.04  & 65.84 & \bf{68.66} & 68.60 & \cellcolor{LightGray!30}70.06 \\
\bottomrule
\end{tabular}}
\caption{Detailed per-language accuracy results on MGSM, XNLI, and MSVAMP for the \llamamodel\ model. Gray dashes indicate that the dataset does not contain that language. B = Baseline, MFS = Multilingual Few-Shot Baseline \cite{tu2025blessing}, OR = Oracle (target-language few-shot reference, not intended for direct comparison). \textbf{Bold} indicates the best score per language excluding the oracle.}
\label{tab:multi_detailed}
\end{table*}
The steering vector $\mathbf{v}(t) \in \mathbb{R}^d$ at layer $t$ is the mean of activation differences:
\begin{equation}
    \mathbf{v}(t) = \frac{1}{N} \sum_{i=1}^{N} \left( h_i^{L_t}(t) - h_i^{L_s}(t) \right)
\end{equation}
\noindent where $N = |\mathcal{D}_{\text{compute}}|$ is the number of samples in the compute set. This setup is designed to shift activations from the source language distribution toward the target language distribution, with the aim of steering the model's internal representations toward target-language processing.


\subsection{Inference-Time Steering}
\label{sec:inference_steer}
During inference, we apply the pre-computed steering vector to hidden states at specific token positions within the input prompt. Given a test example from $\mathcal{D}_{\text{test}}$ with few-shot demonstrations $\mathcal{F}$ in the source language and a target language question $q_{\text{test}}^{L_t}$, the full input prompt is:
\begin{equation}
    \text{prompt} = [\text{system\_instruction}] \oplus \mathcal{F} \oplus q_{\text{test}}^{L_t}
\end{equation}
\noindent where $\mathcal{F} = \{(q_1^{L_s}, c_1^{L_s}, a_1^{L_s}), \ldots, (q_k^{L_s}, c_k^{L_s}, a_k^{L_s})\}$ consists of $k$ few-shot demonstrations with source language questions $q_i^{L_s}$, source language chain-of-thought reasoning $c_i^{L_s}$, and answers $a_i$. 

\paragraph{Steering Positions.} We define a set of token positions $\mathcal{P}$ where steering is applied. We experiment with four steering position configurations (Table \ref{tab:intervention_detailed}): \texttt{on\_fewshot} (all few-shot tokens), \texttt{after\_fewshot} (boundary token only), \texttt{on\_question} (test question tokens only), and \texttt{entire} (all prompt tokens).


\paragraph{Modified Forward Pass.} We modify the hidden states for each position $p \in \mathcal{P}$ during the forward pass at layer~$t$:
\begin{equation}
    \mathbf{h}_p'(t) = \mathbf{h}_p(t) + \alpha \cdot \mathbf{v}(t)
\end{equation}
\noindent where $\mathbf{h}_p(t) \in \mathbb{R}^d$ is the original hidden state at position $p$ and layer $t$, $\alpha \in \mathbb{R}$ is a scaling hyperparameter controlling steering strength, and $\mathbf{h}_p'(t)$ is the steered hidden state. 
The intervention layer, scaling factor, and token positions $(t^*, \alpha^*, P^*)$ are selected using $\mathcal{D}_{\text{val}}$, with performance reported on $\mathcal{D}_{\text{test}}$.




%% file: results_new.tex
\subsection{Setup}


We test on three instruction-tuned models: \llamamodel, \qwenseven, and \qwenfourteen\ \cite{grattafiori2024llama, qwen2025qwen25technicalreport}. We consider three datasets: MGSM (mathematical reasoning) \cite{shi2023language}, MSVAMP (arithmetic word problems) \cite{chen2024breaking}, and XNLI (natural language inference) \cite{conneau2018xnli}.

\paragraph{Data Splits.} Several benchmarks lack a usable training split in all target languages (MGSM provides only 8 exemplars per language and MSVAMP none), so we partition each benchmark's test set into three disjoint parts: $\mathcal{D}_{\text{compute}}$ for vector computation, $\mathcal{D}_{\text{val}}$ for configuration selection, and $\mathcal{D}_{\text{test}}$ for final evaluation. Because the datasets are parallel, each item ID is assigned to the same split in every language, so no $\mathcal{D}_{\text{test}}$ item is seen in any language during vector computation or configuration selection. We apply the same protocol to XNLI for consistency across datasets. To confirm that the gains do not depend on drawing $\mathcal{D}_{\text{compute}}$ and $\mathcal{D}_{\text{val}}$ from the evaluation benchmark, we compute vectors and select configurations entirely on MathOctopus and evaluate on MGSM (Table~\ref{tab:offtest-best-on-test}): steering improves the baseline from 63.5\% to 65.9\%, close to the 67.9\% obtained with in-benchmark data.
 
\paragraph{Implementation Details.}
\begin{table*}[t]
\centering
\setlength{\tabcolsep}{5pt}
\resizebox{\textwidth}{!}{
\begin{tabular}{lcccccccccccc}
\toprule
\multirow{2}{*}{Model} & \multicolumn{4}{c}{MGSM} & \multicolumn{4}{c}{XNLI} & \multicolumn{4}{c}{MSVAMP} \\
\cmidrule(lr){2-5}\cmidrule(lr){6-9}\cmidrule(lr){10-13}
& B & MFS & Ours & \cellcolor{LightGray!30}OR
& B & MFS & Ours & \cellcolor{LightGray!30}OR
& B & MFS & Ours & \cellcolor{LightGray!30}OR \\
\midrule
\llamamodel
& 61.01 & 63.19 & \bf{65.87} & \cellcolor{LightGray!30}68.16
& 61.38 & 61.68 & \bf{64.31} & \cellcolor{LightGray!30}65.04
& 65.84 & \bf{68.66} & {68.60} & \cellcolor{LightGray!30}70.06 \\
\qwenseven
& 68.95 & 69.94 & \bf{71.83} & \cellcolor{LightGray!30}73.21
& 74.81 & 73.27 & \bf{75.51} & \cellcolor{LightGray!30}74.53
& 75.18 & \bf{76.94} & 76.68 & \cellcolor{LightGray!30}77.65 \\
\qwenfourteen
& 81.55 & 82.84 & \bf{84.23} & \cellcolor{LightGray!30}84.72
& 72.73 & 73.91 & \bf{74.66} & \cellcolor{LightGray!30}76.45
& 82.10 & 82.90 & \bf{83.90} & \cellcolor{LightGray!30}83.03 \\
\bottomrule
\end{tabular}}
\caption{Average performance comparison across three datasets and three models. B = Baseline, MFS = Multilingual Few-Shot Baseline \cite{tu2025blessing}, OR = Oracle (target-language few-shot reference, not intended for direct comparison). Detailed per-language results are in Appendix~\ref{app:other_models}.}
\label{tab:main_results}
\end{table*}

We use k = 6 question–answer pairs per sample when computing vectors (Section \ref{2.2}), matching the number of demonstrations at inference.
We perform a grid search over steering layers $t \in \{5, 10, 15, 20, 25, 30\}$, scaling factors $\alpha \in \{0.5, 1.0, 2.0, 3.0\}$, and the four steering position configurations on $\mathcal{D}_{\text{val}}$. The reported results on $\mathcal{D}_{\text{test}}$ correspond to per-language validation-selected configurations. Only configurations that improve over the baseline on $D_{val}$ are carried forward; if none does, the method defaults to the unsteered baseline (e.g., Japanese on MGSM). To show the method does not depend on per-language selection, we also report a single task-level configuration shared across all languages (Appendix \ref{app:hparam}). 

\paragraph{Baselines.} We compare our method against two baselines: (1) \textit{Source baseline (B)}: few-shot prompts with source questions and source chain-of-thought and answer; (2) \textit{Multilingual few-shot (MFS)}: few-shot examples drawn from multiple languages with their respective answers \citep{tu2025blessing}, representing the approach of diversifying few-shot examples across languages. For reference, we also evaluate a \textit{Target-language oracle (OR)}, which uses target-language few-shot demonstrations to assess performance when the demonstration and query languages match. Following \citet{tu2025blessing}, wherever there is a chain of thought reasoning involved, it's always in the source language (English).

\subsection{Results}
\paragraph{Performance on Llama-3.1-8B-Instruct.} Table~\ref{tab:multi_detailed} reports per-language accuracy on $\mathcal{D}_{\text{test}}$. Steering improves average accuracy over English few-shot prompting by 4.86 percentage points on MGSM, 2.93 on XNLI, and 2.76 on MSVAMP, without parameter updates. It improves or matches baseline performance in 11 of 12 MGSM languages, 13 of 14 XNLI languages, and all nine MSVAMP languages. Steering narrows the gap relative to target-language few-shot prompting in several settings, occasionally exceeding this reference condition.
MGSM shows the largest and most consistent gains. On MSVAMP, steering is within 0.1 points of MFS on average, and on XNLI gains vary more across languages, with MFS ahead in some of them.

\paragraph{Statistical Significance.} We evaluate robustness across three random seeds, varying both few-shot demonstration sampling and steering-vector construction (Table \ref{tab:seed_variance}). Across the 12 languages with complete seed coverage, baseline accuracy averages 61.01\% and steered accuracy averages 66.30\%. The mean improvement is 5.29 percentage points across 12 languages and three seeds, with a 95\% confidence interval of [3.04, 7.77] percentage points. We estimate this interval using a hierarchical bootstrap over languages, seeds, and questions.
 
\paragraph{Performance on Other Models.}
Table~\ref{tab:main_results} summarizes average performance across all languages for each 
model and dataset combination. Our method consistently outperforms the source baseline on 
MGSM across all three models, with improvements ranging from 2.7 to 4.9 points, confirming 
that activation steering is especially effective for structured mathematical reasoning. On 
XNLI and MSVAMP, improvements are generally smaller, with MFS remaining competitive 
especially on MSVAMP. We note, however, that MFS requires curating and labeling few-shot examples across multiple target languages at inference time, whereas our method uses a single fixed English few-shot set and an offline-precomputed steering vector computed from parallel examples. Despite this, our method performs comparably to or better than the 
baselines in the majority of cases across models and datasets. The typological clustering of language vectors (Figure \ref{fig:dendogram}) further supports that these vectors capture real linguistic structure. Per-language results for 
\qwenseven\ and \qwenfourteen\ are in Appendix~\ref{app:other_models}.
\paragraph{Layer, Scale, and Position.} The selected layer and $\alpha$ depend on the task (Table~\ref{tab:best_performing_layers}): layer 10 is chosen for 11 of 14 XNLI languages, and layer 5 is most frequent for MGSM and MSVAMP. Every XNLI language selects $\alpha=3.0$, the largest value searched, while MGSM and MSVAMP select lower values. Steering only the demonstration tokens gives the best average accuracy among the four positions (Table \ref{tab:intervention_detailed}), and a single task-level configuration shared across languages still improves over the baseline on all three datasets (Appendix~\ref{app:hparam}).
\begin{table}[t]
\centering
\small
\setlength{\tabcolsep}{3.5pt}
\begin{tabular}{lccccc}
\toprule
 & B & Ours & Nearest & Farthest & Random \\
\midrule
eu & 32.14 & \textbf{36.90} & 29.76\,{\tiny sw} & 27.38\,{\tiny zh} & 33.73 \\
bn & 57.14 & \textbf{61.90} & 58.33\,{\tiny th} & 59.52\,{\tiny zh} & 59.92 \\
ca & 64.29 & \textbf{69.05} & 64.29\,{\tiny gl} & 64.29\,{\tiny zh} & 63.89 \\
zh & 67.86 & \textbf{71.43} & 69.05\,{\tiny ja} & 69.05\,{\tiny bn} & 67.46 \\
fr & 61.90 & \textbf{70.24} & 65.48\,{\tiny es} & 65.48\,{\tiny bn} & 64.29 \\
gl & 64.29 & \textbf{73.81} & 71.43\,{\tiny ca} & 67.86\,{\tiny zh} & 65.48 \\
de & 66.67 & 75.00 & \textbf{76.19}\,{\tiny fr} & 69.05\,{\tiny bn} & 68.65 \\
ja & 55.95 & 55.95 & 58.33\,{\tiny zh} & 58.33\,{\tiny gl} & \textbf{60.32} \\
ru & 71.43 & \textbf{72.62} & 71.43\,{\tiny es} & \textbf{72.62}\,{\tiny bn} & 70.63 \\
es & \textbf{77.38} & 76.19 & 73.81\,{\tiny fr} & 75.00\,{\tiny bn} & 76.19 \\
sw & 55.95 & \textbf{65.48} & 64.29\,{\tiny ca} & 57.14\,{\tiny zh} & 61.51 \\
th & 57.14 & \textbf{61.90} & 60.71\,{\tiny sw} & 59.52\,{\tiny de} & 57.14 \\
\midrule
Avg. & 61.01 & \textbf{65.87} & 63.59 & 62.10 & 62.43 \\
\bottomrule
\end{tabular}
\caption{Steering with another language's vector on MGSM (\llamamodel).
Ours = own vector; Nearest/Farthest = vector of the closest/most distant
language by layer-10 cosine distance (source language in small type);
Random = Gaussian, mean of 3 draws. All vectors rescaled to the target's
own norm. \textbf{Bold}: best per language.}
\label{tab:crosslingual_near_far_random}
\end{table}
\input{cross_task_avg}
\paragraph{Comparison with Self-Translation.}
We compare our method against translating the English demonstrations of every example into the target language, by using the same model for translation. Evaluating MGSM on \llamamodel, steering
reaches comparable accuracy to this self-translation approach (65.87 vs.\
66.07 on average; five wins, five losses, and two ties across twelve
languages; Table~\ref{tab:steering_cost}) while avoiding its cost:
self-translation requires an additional generation pass per query,
averaging 324 extra tokens, whereas steering adds a single precomputed
vector and generates no extra tokens. That these two methods reach similar
accuracy suggests that much of the benefit of explicit translation in
token space can be recovered by an offset in activation space.

%% file: cross_task_avg.tex
\begin{table}[t]
\centering
\setlength{\tabcolsep}{3.5pt}
\small
\begin{tabular}{lccccc}
\toprule
Vector $\rightarrow$ Eval & B & MFS & Ours & CT & \cellcolor{LightGray!30}OR \\
\midrule
MGSM $\rightarrow$ XNLI   & 75.19 & 74.22 & 75.92 & \bf{75.96} & \cellcolor{LightGray!30}75.58 \\
XNLI $\rightarrow$ MGSM   & 73.64 & 74.32 & 76.53 & \bf{77.04} & \cellcolor{LightGray!30}77.55 \\
\midrule
MGSM $\rightarrow$ MSVAMP & 75.18 & \bf{76.94} & 76.68 & 75.18 & \cellcolor{LightGray!30}77.65 \\
MSVAMP $\rightarrow$ MGSM & 71.69 & 73.54 & \bf{74.34} & 72.22 & \cellcolor{LightGray!30}75.79 \\
\midrule
MSVAMP $\rightarrow$ XNLI & 75.19 & 74.25 & \bf{75.92} & 74.04 & \cellcolor{LightGray!30}75.58 \\
XNLI $\rightarrow$ MSVAMP & 76.05 & 77.41 & \bf{77.63} & 75.83 & \cellcolor{LightGray!30}77.50 \\
\bottomrule
\end{tabular}
\caption{Average cross-task transfer accuracy across languages on \qwenseven. Steering vectors are computed on one task and evaluated on another. B = Baseline, MFS = Multilingual Few-Shot Baseline \cite{tu2025blessing}, CT = Cross-Transfer, OR = Oracle (target-language few-shot reference, not intended for direct comparison). \textbf{Bold} indicates the best non-oracle score.}
\label{tab:cross_task_avg}
\end{table}

%% file: ablation.tex
\subsection{Linguistic Geometry}
\begin{figure*}
    \centering
    \includegraphics[width=0.9\linewidth]{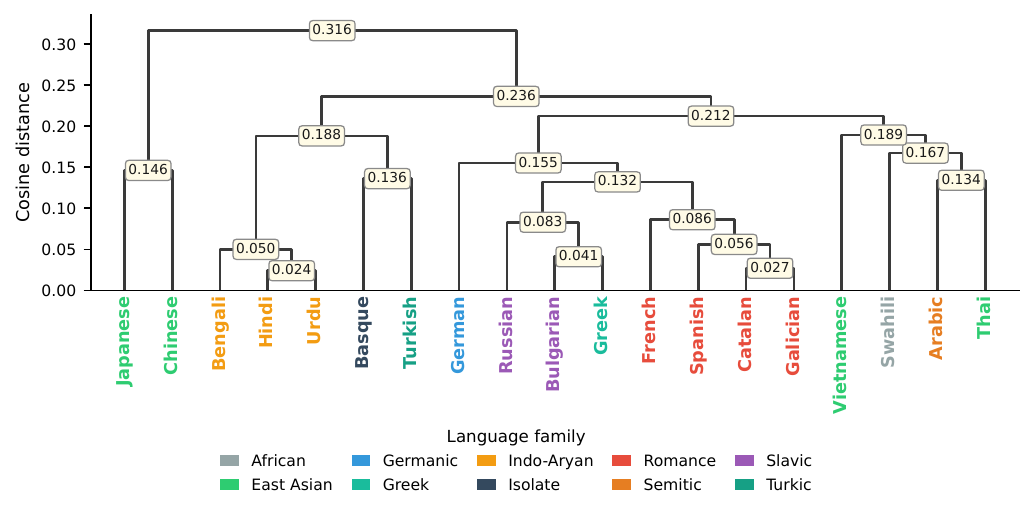}
    \caption{Hierarchical clustering of steering vectors at layer 10 of Llama-3.1-8B-Instruct using the FLORES dataset. 
Edge labels show cosine distances (lower = more similar); colours indicate language families.}
    \label{fig:dendogram}
    \vspace{-1em}
\end{figure*}

If language vectors capture language-associated structure, they should encode known relationships between languages. To test this without confounding by data source, we compute a vector for each of the 19 languages from the same parallel corpus, FLORES-200 \cite{costa2022no}, at layer 10 of Llama-3.1-8B-Instruct, and apply hierarchical clustering with cosine distance as shown in Fig. \ref{fig:dendogram}. 


The closest pairs include Hindi--Urdu
(0.024) and Catalan--Galician (0.027).
Bengali joins the Indo-Aryan cluster
(0.050), while Spanish and French join
the Romance cluster (0.056 and 0.086).
Other groupings include Bulgarian--Greek
(0.041), Basque--Turkish (0.136), and
Japanese--Chinese (0.146), potentially
reflecting areal, typological, or
script-related similarities. The
Arabic--Thai--Swahili--Vietnamese grouping
spans multiple language families, suggesting
that not all clusters reflect genealogical
relatedness.

The structure the vectors recover is thus a mixture of typological relatedness, script, and model-internal similarity, and it is sharpest for the closely related pairs. It is also stable across data sources: vectors computed from task data recover the same closest pairs (Catalan–Galician, Hindi–Urdu, Bulgarian–Greek; Appendix~\ref{app:task_dendogram}). By contrast, random vectors clustered under the same procedure collapse into a near-flat hierarchy, with all languages merging within a narrow band of high distance and no linguistically interpretable substructure (Appendix~\ref{app:random}).

\subsection{Generalization}
If language vectors are reusable, their effects should transfer beyond the data used to build them, so we test transfer across tasks, datasets, and languages.

\paragraph{Across Tasks}
\label{sec:cross_task_transfer}

We evaluate all six transfer directions among MGSM, MSVAMP, and XNLI
on \qwenseven, computing vectors from a source task and applying them
at evaluation on a target task, using the hyperparameters selected for
\textit{Ours} (Table~\ref{tab:cross_task_avg}, per-language results in Appendix \ref{app:cross_task_transfer}.). MGSM and XNLI vectors
transfer to each other's tasks without loss. XNLI vectors reach
$77.04$ on MGSM against a $73.64$ baseline, exceeding the $76.53$
achieved by MGSM's own vectors, and MGSM vectors reach $75.96$ on
XNLI, matching the in-task $75.92$. 
Transfers involving MSVAMP are more variable. These results
suggest that steering effects can
generalize across most tasks.



\paragraph{Across Datasets} 
Beyond task-specific data, we also evaluate vectors computed from general-purpose parallel sentences \citep{costa2022no} and single question-answer pairs; as shown in Table~\ref{tab:multi_flores_raw}, both improve over the source baseline for most configurations, though concatenated few-shot sequences generally outperform both alternatives. We also compute vectors and select configurations on a different math dataset, MathOctopus, and evaluate on MGSM (Table~\ref{tab:offtest-best-on-test}). This leads to an improvement from 63.5\% to 65.9\% on MGSM accuracy.
This provides evidence that steering can remain effective without benchmark-specific vector construction or calibration in this cross-dataset setting.


\begin{table}[t]
\centering
\small
\setlength{\tabcolsep}{5pt}
\begin{tabular}{lrrrrr}
\toprule
Lang. & Baseline & Ours & ST & $\Delta$Tok. & Lat./Q. (s) \\
\midrule
zh & 67.86 & 71.43 & 71.43 & 227 & 12.64 \\
bn & 57.14 & 61.90 & 57.14 & 872 & 26.87 \\
de & 66.67 & 75.00 & 72.62 & 231 & 13.40 \\
es & 77.38 & 76.19 & 80.95 & 244 & 12.25 \\
fr & 61.90 & 70.24 & 67.86 & 264 & 15.14 \\
ja & 55.95 & 55.95 & 57.14 & 310 & 16.97 \\
ru & 71.43 & 72.62 & 67.86 & 267 & 14.07 \\
sw & 55.95 & 65.48 & 60.71 & 315 & 16.00 \\
th & 57.14 & 61.90 & 65.48 & 361 & 18.21 \\
ca & 64.29 & 69.05 & 72.62 & 255 & 14.05 \\
eu & 32.14 & 36.90 & 45.24 & 280 & 14.38 \\
gl & 64.29 & 73.81 & 73.81 & 264 & 13.47 \\
\midrule
Mean & 61.01 & 65.87 & 66.07 & 324 & 15.62 \\
\bottomrule
\end{tabular}
\caption{Accuracy (\%) of baseline, steering, and self-translate (few-shot demos re-translated per query), with self-translate's extra generated tokens ($\Delta$Tok.) and latency per query. Steering adds no extra tokens.}
\label{tab:steering_cost}
\end{table}

\paragraph{Across Languages.} If each vector encodes its own language, replacing it with another language's vector should reduce the gain, and more so for distant languages. We test this on MGSM by steering each language with the vector of its nearest language, its farthest language, or a random direction, each rescaled to the same norm and selected under the same protocol (Table~\ref{tab:crosslingual_near_far_random}). The own-language vector performs best, reaching 65.87 on average against 63.59 for the nearest language, 62.43 for random vectors, and 62.10 for the farthest language, and it is the best or tied-best of all 12 vectors in 8 of 12 languages. On average, nearby language directions in activation space yield higher accuracy than distant directions. These results suggest a relationship between representational similarity and cross-language steering effectiveness.

%% file: related_work.tex
\paragraph{Cross-Lingual Transfer and Alignment.}
Cross-lingual transfer learning investigates how multilingual models transfer knowledge across languages \citep{HuangAPC21, HuangHNCP22, qin2025survey}. To mitigate cross-lingual performance gaps, alignment strategies have been proposed ranging from prompt-level techniques such as multilingual prompting and code-switching, to representation-level methods and cross-lingual in-context pretraining \citep{li2024improving, zhang-etal-2024-plug, wu-etal-2025-enhancing-llm, tu2025blessing, ahuja-etal-2023-mega}. Translation-based approaches offer an alternative: \citet{shi2023language} demonstrate that multilingual chain-of-thought reasoning benefits from English-language reasoning chains, and \citet{liu-etal-2025-translation} and \citet{zhu-etal-2024-question} examine when translating inputs into English yields downstream gains. 
Unlike these approaches, we avoid translation and additional training, steering representations during inference.

\paragraph{Multilingual Representations and Interpretability.}
Multilingual models encode language identity and linguistic structure as structured patterns within their activations \citep{zhao2024large, gurgurov2025language, tang2024language, pokharel2026cross}. \citet{wendler-etal-2024-llamas} and \citet{schut2025multilingual} show that intermediate representations are biased toward English even for non-English inputs, while \citet{dumas-etal-2025-separating} demonstrate via activation patching that concept representations can be separated from language identity, suggesting that language occupies structured, manipulable directions in activation space. These insights motivate activation-level interventions without retraining \citep{nie2025mechanistic}. \citet{pokharel2026cross} propose CLAS, which modulates neuron activations to identify shared and language-specific neurons for zero-shot classification. Interventions on language-specific neurons or sparse dimensions have been used to control output language \citep{kojima-etal-2024-multilingual, zhong-etal-2026-language} and to align cross-lingual representations for retrieval \citep{sundar-etal-2025-steering}, but they are not evaluated on, or reduce, downstream task accuracy; we instead target task accuracy in cross-lingual ICL.
\paragraph{Activation Steering and Language Vectors.}
Activation steering modifies internal activations at inference 
time to control model behavior \citep{turner2023steering}. \citet{zou2023representation} established a top-down framework for reading and writing interpretable directions in representation space; foundational work showed that mean activation differences between contrastive inputs encode such directions \citep{marks2023geometry, rimsky2024steering} and can shift model behavior as inference-time offsets \citep{park2023linear}. \citet{jorgensen2023improving} further showed that mean-centering improves 
steering vector quality,
and \citet{liu2023context} showed that vectors from few-shot English 
demonstrations can steer downstream task behavior, forming the 
methodological precursor to our approach.
More recent work proposes lightweight post-training steering methods, showing that simple residual stream interventions reliably shift model behavior \citep{cui2025painless, stolfo2024improving}. More recent work applies language-specific vectors to language confusion, where models output in an unintended language \citep{yunfan2025mitigating, sterz2025recover, lopo2025surgery}; notably, \citet{sterz2025recover} observe that prior steering approaches tend to harm downstream task accuracy, while \citet{lopo2025surgery} similarly target output language fidelity rather than task accuracy. 

\paragraph{In-Context Learning.}
ICL enables task adaptation without gradient updates  \citep{brown2020language, dong2024survey}. \citet{hendel-etal-2023-context} showed that ICL compresses demonstrations into implicit task vectors, and \citet{todd2024function} extended this to function vectors that generalize across contexts. Unlike these task-encoding approaches, our vectors capture language shift rather than task structure. Cross-lingual ICL studies show that performance degrades when demonstrations are in a high-resource language but test queries are in a low-resource one \citep{tanwar-etal-2023-multilingual, winata2021language}, precisely the gap our method targets. Demonstration-selection methods such as BMF-ICL \citep{kaneko2025balanced}
and XAMPLER \citep{lin2025xampler} improve cross-lingual ICL by changing
which examples the model sees; we instead keep the English demonstrations
fixed and steer their internal representations, an approach complementary
to selection.

%% file: conclusion.tex
We introduce language vectors, a training-free approach
that applies source--target activation differences as
inference-time offsets. Across three datasets, 19
languages, and three models, steering improves
cross-lingual ICL accuracy. The vectors also transfer
across datasets, several task settings, and related
languages, while exhibiting interpretable linguistic
structure. These findings suggest that language-associated
information can be leveraged through activation-space
interventions.

\section{Limitations} 

Our evaluation is limited to instruction-tuned 
decoder-only models. We focus on single-layer interventions with English as the source language. A systematic comparison with multi-layer and mixed-source alternatives remains outside its scope, and we leave it for future work. Understanding why certain task transfers fail, extending the approach to 
languages without parallel data, applying language vectors to long-context tasks such as summarisation remain promising directions for future work.


%% file: appendix.tex
\section{Per-Language Results for \qwenseven\ and \qwenfourteen}
\label{app:other_models}
Tables \ref{tab:main_qwen7b} and \ref{tab:qwen_14b} show the results on the \qwenseven\ and \qwenfourteen\ models for all the datasets. Results demonstrate consistent improvements over the source baseline across most languages and datasets.
\begin{table*}[h]
\centering
\setlength{\tabcolsep}{4pt}
\small
\begin{tabular}{lcccccccccccc}
\toprule
\multirow{2}{*}{Language} & \multicolumn{4}{c}{MGSM} & \multicolumn{4}{c}{XNLI} & \multicolumn{4}{c}{MSVAMP} \\
\cmidrule(lr){2-5}\cmidrule(lr){6-9}\cmidrule(lr){10-13}
& B & MFS & Ours & \cellcolor{LightGray!30}OR
& B & MFS & Ours & \cellcolor{LightGray!30}OR
& B & MFS & Ours & \cellcolor{LightGray!30}OR \\
\midrule

Arabic     & \textcolor{gray}{--} & \textcolor{gray}{--} & \textcolor{gray}{--} & \cellcolor{LightGray!30}\textcolor{gray}{--}
           & 70.96 & 71.26 & \bf{71.56} & \cellcolor{LightGray!30}74.55
           & \textcolor{gray}{--} & \textcolor{gray}{--} & \textcolor{gray}{--} & \cellcolor{LightGray!30}\textcolor{gray}{--} \\

Basque     & 21.43 & 15.48 & \bf{26.19} & \cellcolor{LightGray!30}26.19
           & \textcolor{gray}{--} & \textcolor{gray}{--} & \textcolor{gray}{--} & \cellcolor{LightGray!30}\textcolor{gray}{--}
           & \textcolor{gray}{--} & \textcolor{gray}{--} & \textcolor{gray}{--} & \cellcolor{LightGray!30}\textcolor{gray}{--} \\

Bengali    & 58.33 & \bf{66.67} & 58.33 & \cellcolor{LightGray!30}65.48
           & \textcolor{gray}{--} & \textcolor{gray}{--} & \textcolor{gray}{--} & \cellcolor{LightGray!30}\textcolor{gray}{--}
           & 65.57 & \bf{70.96} & 67.07 & \cellcolor{LightGray!30}76.35 \\

Bulgarian  & \textcolor{gray}{--} & \textcolor{gray}{--} & \textcolor{gray}{--} & \cellcolor{LightGray!30}\textcolor{gray}{--}
           & 79.34 & 76.65 & \bf{80.24} & \cellcolor{LightGray!30}78.74
           & \textcolor{gray}{--} & \textcolor{gray}{--} & \textcolor{gray}{--} & \cellcolor{LightGray!30}\textcolor{gray}{--} \\

Catalan    & 80.95 & 78.57 & \bf{84.52} & \cellcolor{LightGray!30}85.71
           & \textcolor{gray}{--} & \textcolor{gray}{--} & \textcolor{gray}{--} & \cellcolor{LightGray!30}\textcolor{gray}{--}
           & \textcolor{gray}{--} & \textcolor{gray}{--} & \textcolor{gray}{--} & \cellcolor{LightGray!30}\textcolor{gray}{--} \\

Chinese    & 84.52 & \bf{88.10} & 86.90 & \cellcolor{LightGray!30}79.76
           & 78.14 & 77.84 & 77.84 & \cellcolor{LightGray!30}79.34
           & 82.04 & 83.23 & \bf{83.53} & \cellcolor{LightGray!30}82.34 \\

French     & 78.57 & \bf{80.95} & 79.76 & \cellcolor{LightGray!30}84.52
           & 81.44 & 82.34 & 82.34 & \cellcolor{LightGray!30}82.63
           & 82.34 & \bf{85.03} & 84.13 & \cellcolor{LightGray!30}84.43 \\

Galician   & 79.76 & \bf{83.33} & 82.14 & \cellcolor{LightGray!30}84.52
           & \textcolor{gray}{--} & \textcolor{gray}{--} & \textcolor{gray}{--} & \cellcolor{LightGray!30}\textcolor{gray}{--}
           & \textcolor{gray}{--} & \textcolor{gray}{--} & \textcolor{gray}{--} & \cellcolor{LightGray!30}\textcolor{gray}{--} \\

German     & 83.33 & 83.33 & \bf{85.71} & \cellcolor{LightGray!30}86.90
           & 81.74 & 77.54 & \bf{82.93} & \cellcolor{LightGray!30}79.34
           & 79.04 & 79.94 & 80.24 & \cellcolor{LightGray!30}81.14 \\

Greek      & \textcolor{gray}{--} & \textcolor{gray}{--} & \textcolor{gray}{--} & \cellcolor{LightGray!30}\textcolor{gray}{--}
           & 75.45 & 72.16 & \bf{76.95} & \cellcolor{LightGray!30}70.96
           & \textcolor{gray}{--} & \textcolor{gray}{--} & \textcolor{gray}{--} & \cellcolor{LightGray!30}\textcolor{gray}{--} \\

Hindi      & \textcolor{gray}{--} & \textcolor{gray}{--} & \textcolor{gray}{--} & \cellcolor{LightGray!30}\textcolor{gray}{--}
           & 73.95 & 72.75 & \bf{74.85} & \cellcolor{LightGray!30}73.35
           & \textcolor{gray}{--} & \textcolor{gray}{--} & \textcolor{gray}{--} & \cellcolor{LightGray!30}\textcolor{gray}{--} \\

Japanese   & 71.43 & \bf{75.00} & 75.00 & \cellcolor{LightGray!30}73.81
           & \textcolor{gray}{--} & \textcolor{gray}{--} & \textcolor{gray}{--} & \cellcolor{LightGray!30}\textcolor{gray}{--}
           & 78.74 & \bf{79.64} & 79.64 & \cellcolor{LightGray!30}79.94 \\

Russian    & 85.71 & 85.71 & \bf{88.10} & \cellcolor{LightGray!30}85.71
           & 78.14 & 79.04 & \bf{79.64} & \cellcolor{LightGray!30}78.74
           & 79.64 & \bf{82.93} & 81.74 & \cellcolor{LightGray!30}81.44 \\

Spanish    & 84.52 & 83.33 & \bf{89.29} & \cellcolor{LightGray!30}86.90
           & 82.63 & 82.63 & \bf{83.53} & \cellcolor{LightGray!30}81.74
           & 82.04 & 83.53 & \bf{83.83} & \cellcolor{LightGray!30}84.13 \\

Swahili    & 19.05 & 16.67 & \bf{26.19} & \cellcolor{LightGray!30}33.33
           & 53.29 & 47.31 & \bf{55.09} & \cellcolor{LightGray!30}54.79
           & 49.10 & 47.90 & \bf{50.60} & \cellcolor{LightGray!30}50.30 \\

Thai       & 79.76 & \bf{82.14} & 79.76 & \cellcolor{LightGray!30}85.71
           & 70.96 & \bf{73.05} & 70.06 & \cellcolor{LightGray!30}72.46
           & 78.14 & \bf{79.34} & 79.34 & \cellcolor{LightGray!30}78.74 \\

Turkish    & \textcolor{gray}{--} & \textcolor{gray}{--} & \textcolor{gray}{--} & \cellcolor{LightGray!30}\textcolor{gray}{--}
           & 76.35 & 74.55 & \bf{76.35} & \cellcolor{LightGray!30}72.75
           & \textcolor{gray}{--} & \textcolor{gray}{--} & \textcolor{gray}{--} & \cellcolor{LightGray!30}\textcolor{gray}{--} \\

Urdu       & \textcolor{gray}{--} & \textcolor{gray}{--} & \textcolor{gray}{--} & \cellcolor{LightGray!30}\textcolor{gray}{--}
           & 65.57 & 62.87 & \bf{66.17} & \cellcolor{LightGray!30}64.67
           & \textcolor{gray}{--} & \textcolor{gray}{--} & \textcolor{gray}{--} & \cellcolor{LightGray!30}\textcolor{gray}{--} \\

Vietnamese & \textcolor{gray}{--} & \textcolor{gray}{--} & \textcolor{gray}{--} & \cellcolor{LightGray!30}\textcolor{gray}{--}
           & 79.34 & 75.75 & \bf{79.64} & \cellcolor{LightGray!30}79.34
           & \textcolor{gray}{--} & \textcolor{gray}{--} & \textcolor{gray}{--} & \cellcolor{LightGray!30}\textcolor{gray}{--} \\
\midrule
Average & 68.95 & 69.94 & \textbf{71.83} & \cellcolor{LightGray!30}73.21
        & 74.81 & 73.27 & \textbf{75.51} & \cellcolor{LightGray!30}74.53
        & 75.18 & 76.94 & \textbf{76.68} & \cellcolor{LightGray!30}77.65 \\
\bottomrule
\end{tabular}

\caption{Detailed per-language accuracy results on MGSM, XNLI, and MSVAMP on \qwenseven. Gray dashes indicate that the dataset does not contain that language. B = Baseline, MFS = Multilingual Few-Shot Baseline, OR = Oracle (target-language few-shot reference, not intended for direct comparison). \textbf{Bold} indicates the best score per language (excluding Oracle).}
\label{tab:main_qwen7b}
\end{table*}

\begin{table*}[h]
\centering
\setlength{\tabcolsep}{4pt}
\small

\begin{tabular}{lcccccccccccc}
\toprule
\multirow{2}{*}{Language} & \multicolumn{4}{c}{MGSM} & \multicolumn{4}{c}{XNLI} & \multicolumn{4}{c}{MSVAMP} \\
\cmidrule(lr){2-5}\cmidrule(lr){6-9}\cmidrule(lr){10-13}
& B & MFS & Ours & \cellcolor{LightGray!30}OR
& B & MFS & Ours & \cellcolor{LightGray!30}OR
& B & MFS & Ours & \cellcolor{LightGray!30}OR \\
\midrule

Arabic      & \textcolor{gray}{--} & \textcolor{gray}{--} & \textcolor{gray}{--} & \cellcolor{LightGray!30}\textcolor{gray}{--} & 73.35 & \textbf{75.15} & 74.85 & \cellcolor{LightGray!30}78.44 & \textcolor{gray}{--} & \textcolor{gray}{--} & \textcolor{gray}{--} & \cellcolor{LightGray!30}\textcolor{gray}{--} \\

Basque      & 47.62 & \textbf{54.76} & 47.62 & \cellcolor{LightGray!30}59.52 & \textcolor{gray}{--} & \textcolor{gray}{--} & \textcolor{gray}{--} & \cellcolor{LightGray!30}\textcolor{gray}{--} & \textcolor{gray}{--} & \textcolor{gray}{--} & \textcolor{gray}{--} & \cellcolor{LightGray!30}\textcolor{gray}{--} \\

Bengali     & 76.19 & \textbf{80.95} & 76.19 & \cellcolor{LightGray!30}77.38 & \textcolor{gray}{--} & \textcolor{gray}{--} & \textcolor{gray}{--} & \cellcolor{LightGray!30}\textcolor{gray}{--} & 75.45 & \textbf{76.95} & \textbf{76.95} & \cellcolor{LightGray!30}76.05 \\

Bulgarian   & \textcolor{gray}{--} & \textcolor{gray}{--} & \textcolor{gray}{--} & \cellcolor{LightGray!30}\textcolor{gray}{--} & 76.35 & \textbf{78.14} & 77.84 & \cellcolor{LightGray!30}79.64 & \textcolor{gray}{--} & \textcolor{gray}{--} & \textcolor{gray}{--} & \cellcolor{LightGray!30}\textcolor{gray}{--} \\

Catalan     & 84.52 & \textbf{89.29} & \textbf{89.29} & \cellcolor{LightGray!30}94.05 & \textcolor{gray}{--} & \textcolor{gray}{--} & \textcolor{gray}{--} & \cellcolor{LightGray!30}\textcolor{gray}{--} & \textcolor{gray}{--} & \textcolor{gray}{--} & \textcolor{gray}{--} & \cellcolor{LightGray!30}\textcolor{gray}{--} \\

Chinese     & 90.48 & 91.67 & \textbf{94.05} & \cellcolor{LightGray!30}89.29 & 77.25 & 76.05 & \textbf{77.84} & \cellcolor{LightGray!30}78.74 & 86.23 & 84.73 & \textbf{87.72} & \cellcolor{LightGray!30}85.63 \\

French      & 90.48 & \textbf{91.67} & \textbf{91.67} & \cellcolor{LightGray!30}90.48 & 78.44 & 80.24 & \textbf{82.04} & \cellcolor{LightGray!30}82.93 & 85.33 & 86.83 & \textbf{87.43} & \cellcolor{LightGray!30}85.93 \\

Galician    & 84.52 & \textbf{86.90} & \textbf{86.90} & \cellcolor{LightGray!30}92.86 & \textcolor{gray}{--} & \textcolor{gray}{--} & \textcolor{gray}{--} & \cellcolor{LightGray!30}\textcolor{gray}{--} & \textcolor{gray}{--} & \textcolor{gray}{--} & \textcolor{gray}{--} & \cellcolor{LightGray!30}\textcolor{gray}{--} \\

German      & 92.86 & 90.48 & \textbf{96.43} & \cellcolor{LightGray!30}91.67 & 78.74 & 78.74 & \textbf{79.04} & \cellcolor{LightGray!30}81.14 & 85.03 & \textbf{87.13} & 86.83 & \cellcolor{LightGray!30}88.62 \\

Greek       & \textcolor{gray}{--} & \textcolor{gray}{--} & \textcolor{gray}{--} & \cellcolor{LightGray!30}\textcolor{gray}{--} & 72.16 & \textbf{75.45} & \textbf{75.45} & \cellcolor{LightGray!30}77.54 & \textcolor{gray}{--} & \textcolor{gray}{--} & \textcolor{gray}{--} & \cellcolor{LightGray!30}\textcolor{gray}{--} \\

Hindi       & \textcolor{gray}{--} & \textcolor{gray}{--} & \textcolor{gray}{--} & \cellcolor{LightGray!30}\textcolor{gray}{--} & 73.35 & \textbf{75.75} & 74.85 & \cellcolor{LightGray!30}78.14 & \textcolor{gray}{--} & \textcolor{gray}{--} & \textcolor{gray}{--} & \cellcolor{LightGray!30}\textcolor{gray}{--} \\

Japanese    & 88.10 & 86.90 & \textbf{92.86} & \cellcolor{LightGray!30}88.10 & \textcolor{gray}{--} & \textcolor{gray}{--} & \textcolor{gray}{--} & \cellcolor{LightGray!30}\textcolor{gray}{--} & 84.73 & 82.93 & \textbf{85.93} & \cellcolor{LightGray!30}82.93 \\

Russian     & 91.67 & \textbf{95.24} & 94.05 & \cellcolor{LightGray!30}95.24 & 75.15 & \textbf{78.74} & 76.65 & \cellcolor{LightGray!30}79.94 & 83.83 & 85.33 & \textbf{86.23} & \cellcolor{LightGray!30}86.23 \\

Spanish     & 92.86 & 88.10 & \textbf{95.24} & \cellcolor{LightGray!30}91.67 & 79.04 & 78.44 & \textbf{81.44} & \cellcolor{LightGray!30}80.24 & 87.72 & \textbf{89.22} & 88.92 & \cellcolor{LightGray!30}88.32 \\

Swahili     & 50.00 & 51.19 & \textbf{54.76} & \cellcolor{LightGray!30}57.14 & 53.29 & 51.80 & \textbf{54.79} & \cellcolor{LightGray!30}58.08 & 69.46 & 71.26 & \textbf{71.56} & \cellcolor{LightGray!30}73.05 \\

Thai        & 89.29 & 86.90 & \textbf{91.67} & \cellcolor{LightGray!30}89.29 & 72.16 & 72.46 & \textbf{73.05} & \cellcolor{LightGray!30}74.25 & 81.14 & 81.74 & \textbf{83.53} & \cellcolor{LightGray!30}80.54 \\

Turkish     & \textcolor{gray}{--} & \textcolor{gray}{--} & \textcolor{gray}{--} & \cellcolor{LightGray!30}\textcolor{gray}{--} & 71.26 & \textbf{74.85} & 74.55 & \cellcolor{LightGray!30}76.35 & \textcolor{gray}{--} & \textcolor{gray}{--} & \textcolor{gray}{--} & \cellcolor{LightGray!30}\textcolor{gray}{--} \\

Urdu        & \textcolor{gray}{--} & \textcolor{gray}{--} & \textcolor{gray}{--} & \cellcolor{LightGray!30}\textcolor{gray}{--} & 63.77 & 64.07 & \textbf{65.27} & \cellcolor{LightGray!30}67.07 & \textcolor{gray}{--} & \textcolor{gray}{--} & \textcolor{gray}{--} & \cellcolor{LightGray!30}\textcolor{gray}{--} \\

Vietnamese  & \textcolor{gray}{--} & \textcolor{gray}{--} & \textcolor{gray}{--} & \cellcolor{LightGray!30}\textcolor{gray}{--} & 73.95 & 74.85 & \textbf{77.54} & \cellcolor{LightGray!30}77.84 & \textcolor{gray}{--} & \textcolor{gray}{--} & \textcolor{gray}{--} & \cellcolor{LightGray!30}\textcolor{gray}{--} \\

\midrule
Average & 81.55 & 82.84 & \textbf{84.23} & \cellcolor{LightGray!30}84.72
        & 72.73 & 73.91 & \textbf{74.66} & \cellcolor{LightGray!30}76.45
        & 82.10 & 82.90 & \textbf{83.90} & \cellcolor{LightGray!30}83.03 \\
\bottomrule
\end{tabular}
\caption{Detailed per-language accuracy results on MGSM, XNLI, and MSVAMP for the \qwenfourteen\ model. Gray dashes indicate that the dataset does not contain that language. B = Baseline, MFS = Multilingual Few-Shot Baseline, OR = Oracle (target-language few-shot reference, not intended for direct comparison). \textbf{Bold} indicates the best score per language.}
\label{tab:qwen_14b}
\end{table*}

\section{Optimal Layer and Scaling Factor per Language}
\label{appendix:hyperparams}
The optimal steering layer and scaling factor $\alpha$ are both task-dependent.
For XNLI, layer 10 dominates 
(11 out of 14 languages), consistent with middle-layer representations encoding 
the semantic and pragmatic features relevant to natural language inference. For 
MGSM and MSVAMP, layer 5 is most frequent (6/11 and 5/9 languages respectively), 
suggesting that mathematical reasoning benefits from steering at earlier layers 
where surface-level language features are more prominent. The optimal $\alpha$ 
also varies: $\alpha = 3.0$ dominates for XNLI while MGSM and MSVAMP show more 
varied and generally lower optimal values, suggesting that natural language 
inference requires stronger steering than structured reasoning tasks. More details are in Table \ref{tab:best_performing_layers}.

\paragraph{Dataset details.}
Table~\ref{tab:splits} gives the number of examples per language in each split.
For each language, we split the benchmark's test set into three disjoint parts
at random: $\mathcal{D}_\text{compute}$, $\mathcal{D}_\text{val}$ and
$\mathcal{D}_\text{test}$. Because the datasets are parallel, the same item IDs
are assigned to the same split in every language. For XNLI, we first subsample
1{,}000 of the 5{,}010 test examples per language, stratified by label.
Few-shot demonstrations come from MGSM: GSM8K train; XNLI: MNLI/XNLI English
train; MSVAMP: GSM8k train.

\section{Ablation Studies}
\subsection{Random Vector Baseline}
\label{app:random}
To validate that our steering vectors capture meaningful 
language-specific information, we compare against random steering 
vectors generated from a standard normal distribution. The random vectors are explicitly set to norm 1. As shown in 
Table~\ref{tab:random_vs_method2}, our method consistently outperforms 
random steering across all models on the XNLI dataset. 
The absolute gap on XNLI is small (0.19--1.29 points across models), which is expected because the search over layers, scaling factors, and positions can partially adapt even random directions through validation performance. On MGSM, where steering helps most, the gap is larger: own-language vectors reach 65.87 against 62.43 for norm-matched random vectors and outperform them in 10 of 12 languages (Table~\ref{tab:crosslingual_near_far_random}). Table~\ref{tab:random_vs_method2} uses unit-norm random vectors, whereas Table~\ref{tab:crosslingual_near_far_random} rescales them to each language's vector norm.
To further validate that our vectors capture genuine language-specific 
structure, we apply hierarchical clustering to random steering vectors at 
their optimal layers. Unlike language vectors (Figure~\ref{fig:dendogram}), 
random vectors collapse into a near-flat hierarchy with all languages merging within a narrow band of high distance (0.965–1.006, cosine), consistent with near-orthogonal random directions and showing no linguistically meaningful substructure. This confirms that the organized clustering of language identity 
observed in Figure~\ref{fig:dendogram} is not an artifact of hyperparameter 
optimization but reflects language-specific information encoded in 
our computed vectors.

\begin{table}[H]
\centering
\setlength{\tabcolsep}{6pt}
\small
\begin{tabular}{llcc}
\toprule
\multirow{2}{*}{Dataset} & \multirow{2}{*}{Model} & \multicolumn{2}{c}{Performance} \\
\cmidrule(lr){3-4}
& & R & Ours \\
\midrule
\multirow{3}{*}{XNLI}
 & \llamamodel & 63.37 & \bf{64.31} \\
 & \qwenseven   & 75.32 & \bf{75.51} \\
 & \qwenfourteen  & 73.37 & \bf{74.66} \\
\bottomrule
\end{tabular}
\caption{Comparison of random steering vs. our method across models. R = Random Steering.}
\label{tab:random_vs_method2}
\end{table}

\subsection{Steering Vector Ablations: Data Source and Format}
\label{app:flores}

Table \ref{tab:multi_flores_raw} shows the accuracy when general non-task specific data or single question-answer pairs are used to compute the language vector. Single QnA uses the same approach as our method but computes vectors from individual question-answer pairs rather than concatenated few-shot sequences, serving as an ablation of our few-shot format for vector computation. Results show that vectors computed from concatenated few-shot QA sequences consistently outperform both FLORES-based vectors and single in-domain QA-pair vectors, suggesting that multi-example contextual composition provides a stronger steering signal than isolated examples or general-domain translation pairs.
\begin{table}[ht]
\centering
\small
\begin{tabular}{lccc}
\toprule
\textbf{Language} & \textbf{Base} & \textbf{Mean} & \textbf{Last} \\
\midrule
Bengali    & 57.14 & 61.90 & 53.57 \\
Catalan    & 64.29 & 69.05 & 61.90 \\
German     & 66.67 & 75.00 & 65.48 \\
Spanish    & 77.38 & 76.19 & 71.43 \\
Basque     & 32.14 & 36.90 & 32.14 \\
French     & 61.90 & 70.24 & 60.71 \\
Galician   & 64.29 & 73.81 & 67.86 \\
Japanese   & 55.95 & 55.95 & 55.95 \\
Russian    & 71.43 & 72.62 & 66.67 \\
Swahili    & 55.95 & 65.48 & 59.52 \\
Thai       & 57.14 & 61.90 & 57.14 \\
Chinese    & 67.86 & 71.43 & 66.67 \\
\midrule
\textbf{Avg} & 61.01 & 65.87 & 59.92 \\
\bottomrule
\end{tabular}
\caption{Accuracy using mean-pooling vs.\ last-token hidden states (Llama-3.1-8B-Instruct, MGSM). Mean = Mean, Last = Last Token.}
\label{tab:steering_comparison}
\end{table}
\subsection{Token Aggregation: Mean Pooling vs. Last Token}
\label{appendix:pooling}
In Table \ref{tab:steering_comparison} we compare mean-pooling against last-token extraction for the hidden state aggregation when computing steering vectors. Using mean tokens leads to a better performance as compared to just using the last token. Mean pooling captures language-level information distributed across the full sequence, whereas last-token extraction is sensitive to positional artifacts and punctuation tokens at sequence boundaries, which may not carry meaningful language-specific signal. This is consistent with our use of mean pooling for steering vector computation, where the goal is to capture aggregate language identity rather than position-specific features.


\subsection{Steering Positions}
Table \ref{tab:intervention_detailed} reports MGSM accuracy for each steering position configuration on Llama-3.1-8B-Instruct. \texttt{on\_fewshot} achieves the highest average accuracy, although the best position varies by language.
\begin{table}[t]
\centering
\setlength{\tabcolsep}{7pt}
\resizebox{\columnwidth}{!}{
\begin{tabular}{lccccc}
\toprule
Language & B & OF & AF & OQ & ENT \\
\midrule
Bengali & 57.14 & 60.71 & \bf{61.90} & 60.71 & 55.95 \\
Catalan & 64.29 & 66.67 & \bf{69.05} & 67.86 & \bf{69.05} \\
German & 66.67 & \bf{75.00} & 66.67$^\dagger$ & 69.05 & 64.29 \\
Spanish & \bf{77.38} & 73.81 & 75.00 & 76.19 & 72.62 \\
Basque & 32.14 & 35.71 & \bf{36.90} & 33.33 & \bf{36.90} \\
French & 61.90 & 67.86 & 64.29 & 67.86 & \bf{70.24} \\
Galician & 64.29 & \bf{73.81} & 69.05 & 66.67 & \bf{73.81} \\
Japanese & \bf{55.95} & \bf{55.95}$^\dagger$ & \bf{55.95} & \bf{55.95}$^\dagger$ & \bf{55.95}$^\dagger$ \\
Russian & 71.43 & 69.05 & \bf{72.62} & \bf{72.62} & 69.05 \\
Swahili & 55.95 & \bf{65.48} & 60.71 & 63.10 & \bf{65.48} \\
Thai & 57.14 & \bf{61.90} & 54.76 & \bf{61.90} & 59.52 \\
Chinese & 67.86 & \bf{71.43} & \bf{71.43} & \bf{71.43} & \bf{71.43} \\
\midrule
Average & 61.01 & \textbf{64.78} & 63.19 & 63.89 & 63.69 \\
\bottomrule
\end{tabular}}
\caption{Per-language accuracy across steering position configurations for \llamamodel\ on MGSM. B = Baseline, OF = On Few-shot, AF = After Few-shot, OQ = On Question, ENT = Entire Prompt. $^\dagger$No configuration improved over the baseline on validation, so the baseline score is reported. \textbf{Bold} indicates the best score per language.}
\label{tab:intervention_detailed}
\end{table}

\subsection{Cross-Task Transfer}
\label{app:cross_task_transfer}
Detailed results per language for all the transfers are shown in Table \ref{tab:cross_task_corrected}.

\input{cross_task_new}

\label{app:norms}

\begin{table*}[h]
\centering
\setlength{\tabcolsep}{2.5pt}
\small
\resizebox{\textwidth}{!}{
\begin{tabular}{lccccccccccccccc}
\toprule
\multirow{2}{*}{Language} & \multicolumn{5}{c}{MGSM} & \multicolumn{5}{c}{XNLI} & \multicolumn{5}{c}{MSVAMP} \\
\cmidrule(lr){2-6}\cmidrule(lr){7-11}\cmidrule(lr){12-16}
& B & FLORES & Single & Ours & \cellcolor{LightGray!30}OR 
& B & FLORES & Single & Ours & \cellcolor{LightGray!30}OR 
& B & FLORES & Single & Ours & \cellcolor{LightGray!30}OR \\
\midrule
Arabic      & \textcolor{gray}{--} & \textcolor{gray}{--} & \textcolor{gray}{--} & \textcolor{gray}{--} & \cellcolor{LightGray!30}\textcolor{gray}{--}
            & 62.57 & 61.08 & \bf{63.17} & 62.87 & \cellcolor{LightGray!30}63.77
            & \textcolor{gray}{--} & \textcolor{gray}{--} & \textcolor{gray}{--} & \textcolor{gray}{--} & \cellcolor{LightGray!30}\textcolor{gray}{--} \\

Basque      & 32.14 & 27.38 & \bf{36.90} & \bf{36.90} & \cellcolor{LightGray!30}52.38
            & \textcolor{gray}{--} & \textcolor{gray}{--} & \textcolor{gray}{--} & \textcolor{gray}{--} & \cellcolor{LightGray!30}\textcolor{gray}{--}
            & \textcolor{gray}{--} & \textcolor{gray}{--} & \textcolor{gray}{--} & \textcolor{gray}{--} & \cellcolor{LightGray!30}\textcolor{gray}{--} \\

Bengali     & 57.14 & 57.14 & 60.71 & \bf{61.90} & \cellcolor{LightGray!30}58.33
            & \textcolor{gray}{--} & \textcolor{gray}{--} & \textcolor{gray}{--} & \textcolor{gray}{--} & \cellcolor{LightGray!30}\textcolor{gray}{--}
            & 57.49 & 57.49 & 59.28 & \bf{59.58} & \cellcolor{LightGray!30}61.38 \\

Bulgarian   & \textcolor{gray}{--} & \textcolor{gray}{--} & \textcolor{gray}{--} & \textcolor{gray}{--} & \cellcolor{LightGray!30}\textcolor{gray}{--}
            & 56.29 & 58.68 & 60.18 & \bf{61.68} & \cellcolor{LightGray!30}66.17
            & \textcolor{gray}{--} & \textcolor{gray}{--} & \textcolor{gray}{--} & \textcolor{gray}{--} & \cellcolor{LightGray!30}\textcolor{gray}{--} \\

Catalan     & 64.29 & 65.48 & 66.67 & \bf{69.05} & \cellcolor{LightGray!30}76.19
            & \textcolor{gray}{--} & \textcolor{gray}{--} & \textcolor{gray}{--} & \textcolor{gray}{--} & \cellcolor{LightGray!30}\textcolor{gray}{--}
            & \textcolor{gray}{--} & \textcolor{gray}{--} & \textcolor{gray}{--} & \textcolor{gray}{--} & \cellcolor{LightGray!30}\textcolor{gray}{--} \\

Chinese     & 67.86 & 65.48 & \bf{73.81} & 71.43 & \cellcolor{LightGray!30}72.62
            & 59.88 & 56.29 & \bf{60.18} & 59.28 & \cellcolor{LightGray!30}61.98
            & 69.76 & 69.76 & \bf{71.26} & \bf{71.26} & \cellcolor{LightGray!30}73.35 \\

French      & 61.90 & 64.29 & 67.86 & \bf{70.24} & \cellcolor{LightGray!30}65.48
            & 67.37 & 71.26 & 68.86 & \bf{72.75} & \cellcolor{LightGray!30}71.26
            & 71.56 & 73.35 & 73.35 & \bf{74.55} & \cellcolor{LightGray!30}73.05 \\

Galician    & 64.29 & \bf{75.00} & 73.81 & 73.81 & \cellcolor{LightGray!30}77.38
            & \textcolor{gray}{--} & \textcolor{gray}{--} & \textcolor{gray}{--} & \textcolor{gray}{--} & \cellcolor{LightGray!30}\textcolor{gray}{--}
            & \textcolor{gray}{--} & \textcolor{gray}{--} & \textcolor{gray}{--} & \textcolor{gray}{--} & \cellcolor{LightGray!30}\textcolor{gray}{--} \\

German      & 66.67 & 65.48 & 70.24 & \bf{75.00} & \cellcolor{LightGray!30}71.43
            & 64.07 & 57.19 & \bf{67.66} & 66.47 & \cellcolor{LightGray!30}65.27
            & \bf{71.26} & 70.06 & \bf{71.26} & \bf{71.26} & \cellcolor{LightGray!30}76.95 \\

Greek       & \textcolor{gray}{--} & \textcolor{gray}{--} & \textcolor{gray}{--} & \textcolor{gray}{--} & \cellcolor{LightGray!30}\textcolor{gray}{--}
            & 68.26 & 60.18 & \bf{70.06} & \bf{70.06} & \cellcolor{LightGray!30}72.75
            & \textcolor{gray}{--} & \textcolor{gray}{--} & \textcolor{gray}{--} & \textcolor{gray}{--} & \cellcolor{LightGray!30}\textcolor{gray}{--} \\

Hindi       & \textcolor{gray}{--} & \textcolor{gray}{--} & \textcolor{gray}{--} & \textcolor{gray}{--} & \cellcolor{LightGray!30}\textcolor{gray}{--}
            & 61.38 & 60.78 & 64.67 & \bf{64.97} & \cellcolor{LightGray!30}61.98
            & \textcolor{gray}{--} & \textcolor{gray}{--} & \textcolor{gray}{--} & \textcolor{gray}{--} & \cellcolor{LightGray!30}\textcolor{gray}{--} \\

Japanese    & 55.95 & 60.71 & \bf{63.10} & 55.95 & \cellcolor{LightGray!30}63.10
            & \textcolor{gray}{--} & \textcolor{gray}{--} & \textcolor{gray}{--} & \textcolor{gray}{--} & \cellcolor{LightGray!30}\textcolor{gray}{--}
            & 63.17 & 63.17 & 65.57 & \bf{67.96} & \cellcolor{LightGray!30}70.06 \\

Russian     & 71.43 & 67.86 & \bf{72.62} & \bf{72.62} & \cellcolor{LightGray!30}76.19
            & 58.98 & 53.59 & \bf{64.97} & 63.77 & \cellcolor{LightGray!30}60.78
            & 68.86 & 71.56 & \bf{72.16} & \bf{72.16} & \cellcolor{LightGray!30}71.56 \\

Spanish     & \bf{77.38} & 71.43 & 76.19 & 76.19 & \cellcolor{LightGray!30}78.57
            & 66.77 & \bf{70.66} & 69.16 & \bf{70.66} & \cellcolor{LightGray!30}67.37
            & 74.55 & 73.95 & 75.45 & \bf{75.75} & \cellcolor{LightGray!30}74.55 \\

Swahili     & 55.95 & 61.90 & 63.10 & \bf{65.48} & \cellcolor{LightGray!30}66.67
            & 52.40 & 47.01 & \bf{61.08} & 55.99 & \cellcolor{LightGray!30}56.89
            & 56.29 & 58.38 & 57.49 & \bf{60.78} & \cellcolor{LightGray!30}62.87 \\

Thai        & 57.14 & 61.90 & \bf{63.10} & 61.90 & \cellcolor{LightGray!30}59.52
            & 59.88 & 51.20 & \bf{62.87} & 60.78 & \cellcolor{LightGray!30}70.66
            & 59.58 & 61.08 & 63.77 & \bf{64.07} & \cellcolor{LightGray!30}66.77 \\

Turkish     & \textcolor{gray}{--} & \textcolor{gray}{--} & \textcolor{gray}{--} & \textcolor{gray}{--} & \cellcolor{LightGray!30}\textcolor{gray}{--}
            & 62.28 & 64.67 & \bf{66.17} & 65.87 & \cellcolor{LightGray!30}65.57
            & \textcolor{gray}{--} & \textcolor{gray}{--} & \textcolor{gray}{--} & \textcolor{gray}{--} & \cellcolor{LightGray!30}\textcolor{gray}{--} \\

Urdu        & \textcolor{gray}{--} & \textcolor{gray}{--} & \textcolor{gray}{--} & \textcolor{gray}{--} & \cellcolor{LightGray!30}\textcolor{gray}{--}
            & 55.09 & 48.20 & \bf{56.89} & 56.29 & \cellcolor{LightGray!30}57.19
            & \textcolor{gray}{--} & \textcolor{gray}{--} & \textcolor{gray}{--} & \textcolor{gray}{--} & \cellcolor{LightGray!30}\textcolor{gray}{--} \\

Vietnamese  & \textcolor{gray}{--} & \textcolor{gray}{--} & \textcolor{gray}{--} & \textcolor{gray}{--} & \cellcolor{LightGray!30}\textcolor{gray}{--}
            & 64.07 & 63.47 & 67.96 & \bf{68.86} & \cellcolor{LightGray!30}68.86
            & \textcolor{gray}{--} & \textcolor{gray}{--} & \textcolor{gray}{--} & \textcolor{gray}{--} & \cellcolor{LightGray!30}\textcolor{gray}{--} \\

\midrule
Average     & 61.01 & 62.00 & 65.68 & \bf{65.87} & \cellcolor{LightGray!30}68.16
            & 61.38 & 58.88 & \bf{64.56} & 64.31 & \cellcolor{LightGray!30}65.04
            & 65.84 & 66.53 & 67.73 & \bf{68.60} & \cellcolor{LightGray!30}70.06 \\
\bottomrule

\end{tabular}}
\caption{Alternative vector configurations per language. B = Baseline, FLORES = vectors computed from general-purpose parallel sentences \citep{costa2022no}, Single = our method with single question-answer pairs instead of concatenated few-shot sequences, Ours = our full few-shot method, OR = Oracle (target-language few-shot reference, not intended for direct comparison). Bold indicates best among B, FLORES, Single, and Ours.}
\label{tab:multi_flores_raw}
\end{table*}

\begin{table}[t]
\centering
\small
\begin{tabular}{lccc}
\toprule
Language & Baseline & Ours & Other-data \\
\midrule
Chinese  & 67.9 & 71.4 & 69.0 \\
Thai     & 57.1 & 61.9 & 58.3 \\
Bengali  & 57.1 & 61.9 & 56.0 \\
German   & 66.7 & 75.0 & 76.2 \\
Spanish  & 77.4 & 76.2 & 75.0 \\
French   & 61.9 & 70.2 & 66.7 \\
Japanese & 56.0 & 56.0 & 56.0 \\
Russian  & 71.4 & 72.6 & 69.0 \\
Swahili  & 56.0 & 65.5 & 66.7 \\
\midrule
Average  & 63.5 & 67.9 & 65.9 \\
\bottomrule
\end{tabular}
\caption{MGSM accuracy on \llamamodel\ by using data for vector computation and validation from a different math dataset (MathOctopus) and testing on MGSM. We use the subset of 9 languages common to both the datasets.}
\label{tab:offtest-best-on-test}
\end{table}


\section{Hyperparameter Configuration Comparison}
\label{app:hparam}

\begin{table}[h]
\centering\small
\begin{tabular}{@{}lcccc@{}}
\toprule
 & Total & Compute & Val & Test \\
\midrule
MGSM   & 250   & 83  & 83  & 84  \\
XNLI   & 1{,}000 & 333 & 333 & 334 \\
MSVAMP & 1{,}000 & 333 & 333 & 334 \\
\bottomrule
\end{tabular}
\caption{Examples per language in each split.}
\label{tab:splits}
\end{table}

We perform a grid search over steering layers $t \in \{5, 10, 15, 20, 25, 30\}$, 
scaling factors $\alpha \in \{0.5, 1.0, 2.0, 3.0\}$, and four steering position 
configurations (\texttt{on\_fewshot}, \texttt{after\_fewshot}, \texttt{on\_question}, 
\texttt{entire}) on the validation set $\mathcal{D}_\text{val}$. Only configurations 
achieving validation accuracy above the source baseline are carried forward to 
evaluation on $\mathcal{D}_\text{test}$.

To assess the impact of per-language hyperparameter selection, we compare two 
configuration levels: (1) \textbf{Per-language}: each language uses its own optimal 
layer, scaling factor, and steering position; (2) \textbf{Task-level}: a single 
configuration is shared across all languages within a task, selected as the one 
achieving the highest average validation accuracy. Results under both protocols are 
reported in Table~\ref{tab:multi_detailed_hparam}. Task-level configurations remain 
consistently above the source baseline across all datasets, confirming that while per-language selection provides additional gains, the method does not rely on it.

The task-level optimal configurations are: for MGSM, layer 5 with $\alpha = 1.0$ and \texttt{on\_fewshot}; for XNLI, layer 10 with $\alpha = 3.0$ and \texttt{on\_fewshot}; for MSVAMP, layer 5 with $\alpha = 0.5$ and \texttt{on\_fewshot}.
\begin{table}[ht]
\centering
\small
\setlength{\tabcolsep}{4pt}
\begin{tabular}{lccc}
\toprule
\textbf{Language} & \textbf{MGSM} & \textbf{MSVAMP} & \textbf{XNLI} \\
\midrule
Arabic     & ---    & ---    & 10/3.0 \\
Basque     & 5/1.0  & ---    & ---    \\
Bengali    & 20/1.0 & 5/0.5  & ---    \\
Bulgarian  & ---    & ---    & 10/3.0 \\
Catalan    & 20/3.0 & ---    & ---    \\
Chinese    & 30/0.5 & 5/0.5  & 5/3.0  \\
French     & 5/1.0  & 15/0.5 & 10/3.0 \\
Galician   & 5/1.0  & ---    & ---    \\
German     & 5/2.0  & 30/1.0 & 5/3.0  \\
Greek      & ---    & ---    & 10/3.0 \\
Hindi      & ---    & ---    & 10/3.0 \\
Japanese   & ---  & 10/0.5 & ---    \\
Russian    & 30/0.5 & 25/0.5 & 10/3.0 \\
Spanish    & 30/1.0 & 5/0.5  & 10/3.0 \\
Swahili    & 5/1.0  & 5/0.5  & 5/3.0  \\
Thai       & 5/0.5  & 5/2.0  & 10/3.0 \\
Turkish    & ---    & ---    & 10/3.0 \\
Urdu       & ---    & ---    & 10/3.0 \\
Vietnamese & ---    & ---    & 10/3.0 \\
\bottomrule
\end{tabular}
\caption{Best layer/$\alpha$ per language and task for \llamamodel. Each cell shows \textit{layer}/$\alpha$; {---} indicates the language was not evaluated on that task.}
\label{tab:best_performing_layers}
\end{table}

\begin{table*}[ht]
\centering
\setlength{\tabcolsep}{2pt}
\small
\resizebox{\textwidth}{!}{
\begin{tabular}{lcccccccccccccccc}
\toprule
\multirow{2}{*}{Language} & \multicolumn{5}{c}{MGSM} & \multicolumn{5}{c}{XNLI} & \multicolumn{5}{c}{MSVAMP} \\
\cmidrule(lr){2-6}\cmidrule(lr){7-11}\cmidrule(lr){12-16}
& B & MFS & Per-Lang & Task-Lvl & \cellcolor{LightGray!30}OR
& B & MFS & Per-Lang & Task-Lvl & \cellcolor{LightGray!30}OR
& B & MFS & Per-Lang & Task-Lvl & \cellcolor{LightGray!30}OR \\
\midrule
Arabic      & \textcolor{gray}{--} & \textcolor{gray}{--} & \textcolor{gray}{--} & \textcolor{gray}{--} & \cellcolor{LightGray!30}\textcolor{gray}{--} & 62.57 & 60.78 & \bf{62.87} & 61.38 & \cellcolor{LightGray!30}63.77 & \textcolor{gray}{--} & \textcolor{gray}{--} & \textcolor{gray}{--} & \textcolor{gray}{--} & \cellcolor{LightGray!30}\textcolor{gray}{--} \\
Basque      & 32.14 & 35.70 & \bf{36.90} & 32.14 & \cellcolor{LightGray!30}52.38 & \textcolor{gray}{--} & \textcolor{gray}{--} & \textcolor{gray}{--} & \textcolor{gray}{--} & \cellcolor{LightGray!30}\textcolor{gray}{--} & \textcolor{gray}{--} & \textcolor{gray}{--} & \textcolor{gray}{--} & \textcolor{gray}{--} & \cellcolor{LightGray!30}\textcolor{gray}{--} \\
Bengali     & 57.14 & 55.95 & \bf{61.90} & 53.57 & \cellcolor{LightGray!30}58.33 & \textcolor{gray}{--} & \textcolor{gray}{--} & \textcolor{gray}{--} & \textcolor{gray}{--} & \cellcolor{LightGray!30}\textcolor{gray}{--} & 57.49 & \bf{62.87} & 59.58 & 59.58 & \cellcolor{LightGray!30}61.38 \\
Bulgarian   & \textcolor{gray}{--} & \textcolor{gray}{--} & \textcolor{gray}{--} & \textcolor{gray}{--} & \cellcolor{LightGray!30}\textcolor{gray}{--} & 56.29 & \bf{61.98} & 61.68 & 60.18 & \cellcolor{LightGray!30}66.17 & \textcolor{gray}{--} & \textcolor{gray}{--} & \textcolor{gray}{--} & \textcolor{gray}{--} & \cellcolor{LightGray!30}\textcolor{gray}{--} \\
Catalan     & 64.29 & 64.28 & \bf{69.05} & 63.10 & \cellcolor{LightGray!30}76.19 & \textcolor{gray}{--} & \textcolor{gray}{--} & \textcolor{gray}{--} & \textcolor{gray}{--} & \cellcolor{LightGray!30}\textcolor{gray}{--} & \textcolor{gray}{--} & \textcolor{gray}{--} & \textcolor{gray}{--} & \textcolor{gray}{--} & \cellcolor{LightGray!30}\textcolor{gray}{--} \\
Chinese     & 67.86 & 67.86 & \bf{71.43} & 70.24 & \cellcolor{LightGray!30}72.62 & 59.88 & \bf{61.38} & 59.28 & 60.78 & \cellcolor{LightGray!30}61.98 & 69.76 & \bf{73.05} & 71.26 & 71.26 & \cellcolor{LightGray!30}73.35 \\
French      & 61.90 & 64.29 & \bf{70.24} & 66.67 & \cellcolor{LightGray!30}65.48 & 67.37 & 68.26 & \bf{72.75} & 68.86 & \cellcolor{LightGray!30}71.26 & 71.56 & 73.05 & \bf{74.55} & 73.65 & \cellcolor{LightGray!30}73.05 \\
Galician    & 64.29 & 69.04 & \bf{73.81} & \bf{73.81} & \cellcolor{LightGray!30}77.38 & \textcolor{gray}{--} & \textcolor{gray}{--} & \textcolor{gray}{--} & \textcolor{gray}{--} & \cellcolor{LightGray!30}\textcolor{gray}{--} & \textcolor{gray}{--} & \textcolor{gray}{--} & \textcolor{gray}{--} & \textcolor{gray}{--} & \cellcolor{LightGray!30}\textcolor{gray}{--} \\
German      & 66.67 & 66.67 & \bf{75.00} & 70.24 & \cellcolor{LightGray!30}71.43 & 64.07 & 65.27 & \bf{66.47} & 65.27 & \cellcolor{LightGray!30}65.27 & 71.26 & 70.06 & \bf{71.26} & 69.16 & \cellcolor{LightGray!30}76.95 \\
Greek       & \textcolor{gray}{--} & \textcolor{gray}{--} & \textcolor{gray}{--} & \textcolor{gray}{--} & \cellcolor{LightGray!30}\textcolor{gray}{--} & 68.26 & 66.17 & \bf{70.06} & \bf{70.06} & \cellcolor{LightGray!30}72.75 & \textcolor{gray}{--} & \textcolor{gray}{--} & \textcolor{gray}{--} & \textcolor{gray}{--} & \cellcolor{LightGray!30}\textcolor{gray}{--} \\
Hindi       & \textcolor{gray}{--} & \textcolor{gray}{--} & \textcolor{gray}{--} & \textcolor{gray}{--} & \cellcolor{LightGray!30}\textcolor{gray}{--} & 61.38 & 57.78 & \bf{64.97} & 64.37 & \cellcolor{LightGray!30}61.98 & \textcolor{gray}{--} & \textcolor{gray}{--} & \textcolor{gray}{--} & \textcolor{gray}{--} & \cellcolor{LightGray!30}\textcolor{gray}{--} \\
Japanese    & 55.95 & \bf{61.90} & 55.95 & \bf{61.90} & \cellcolor{LightGray!30}63.10 & \textcolor{gray}{--} & \textcolor{gray}{--} & \textcolor{gray}{--} & \textcolor{gray}{--} & \cellcolor{LightGray!30}\textcolor{gray}{--} & 63.17 & \bf{68.26} & 67.96 & 67.66 & \cellcolor{LightGray!30}70.06 \\
Russian     & 71.43 & 71.43 & \bf{72.62} & 67.86 & \cellcolor{LightGray!30}76.19 & 58.98 & \bf{64.37} & 63.77 & 60.48 & \cellcolor{LightGray!30}60.78 & 68.86 & \bf{72.46} & 72.16 & 70.66 & \cellcolor{LightGray!30}71.56 \\
Spanish     & \bf{77.38} & 70.24 & 76.19 & 70.24 & \cellcolor{LightGray!30}78.57 & 66.77 & 66.17 & \bf{70.66} & 68.86 & \cellcolor{LightGray!30}67.37 & 74.55 & 73.95 & \bf{75.75} & \bf{75.75} & \cellcolor{LightGray!30}74.55 \\
Swahili     & 55.95 & 63.10 & \bf{65.48} & \bf{65.48} & \cellcolor{LightGray!30}66.67 & 52.40 & 51.20 & \bf{55.99} & 55.09 & \cellcolor{LightGray!30}56.89 & 56.29 & 59.58 & \bf{60.78} & 57.49 & \cellcolor{LightGray!30}62.87 \\
Thai        & 57.14 & \bf{67.86} & 61.90 & 60.71 & \cellcolor{LightGray!30}59.52 & 59.88 & \bf{64.37} & 60.78 & 60.78 & \cellcolor{LightGray!30}70.66 & 59.58 & \bf{64.67} & 64.07 & 61.98 & \cellcolor{LightGray!30}66.77 \\
Turkish     & \textcolor{gray}{--} & \textcolor{gray}{--} & \textcolor{gray}{--} & \textcolor{gray}{--} & \cellcolor{LightGray!30}\textcolor{gray}{--} & 62.28 & 57.19 & \bf{65.87} & 62.87 & \cellcolor{LightGray!30}65.57 & \textcolor{gray}{--} & \textcolor{gray}{--} & \textcolor{gray}{--} & \textcolor{gray}{--} & \cellcolor{LightGray!30}\textcolor{gray}{--} \\
Urdu        & \textcolor{gray}{--} & \textcolor{gray}{--} & \textcolor{gray}{--} & \textcolor{gray}{--} & \cellcolor{LightGray!30}\textcolor{gray}{--} & 55.09 & 55.39 & \bf{56.29} & 54.49 & \cellcolor{LightGray!30}57.19 & \textcolor{gray}{--} & \textcolor{gray}{--} & \textcolor{gray}{--} & \textcolor{gray}{--} & \cellcolor{LightGray!30}\textcolor{gray}{--} \\
Vietnamese  & \textcolor{gray}{--} & \textcolor{gray}{--} & \textcolor{gray}{--} & \textcolor{gray}{--} & \cellcolor{LightGray!30}\textcolor{gray}{--} & 64.07 & 63.17 & \bf{68.86} & 66.77 & \cellcolor{LightGray!30}68.86 & \textcolor{gray}{--} & \textcolor{gray}{--} & \textcolor{gray}{--} & \textcolor{gray}{--} & \cellcolor{LightGray!30}\textcolor{gray}{--} \\
\midrule
Average & 61.01 & 63.19 & \bf{65.87} & 63.00 & \cellcolor{LightGray!30}68.16 & 61.38 & 61.68 & \bf{64.31} & 62.87 & \cellcolor{LightGray!30}65.04 & 65.84 & \bf{68.66} & 68.60 & 67.47 & \cellcolor{LightGray!30}70.06 \\
\bottomrule
\end{tabular}}
\caption{Per-language accuracy under per-language and task-level hyperparameter selection on MGSM, XNLI, and MSVAMP for the \llamamodel\ model. Gray dashes indicate that the dataset does not contain that language. B = Baseline, MFS = Multilingual Few-Shot Baseline \cite{tu2025blessing}, OR = Oracle (target-language few-shot reference, not intended for direct comparison). \textbf{Bold} indicates the best score per language excluding the oracle.}
\label{tab:multi_detailed_hparam}
\end{table*}



\section{Performance Across Seeds and Significance}
We run our steering experiment three times using three different seeds (42, 24, 25), which vary both the few-shot demonstration sampling and the questions used to compute the steering vectors. Table \ref{tab:seed_variance} reports mean ± standard deviation across seeds for the baseline, steered ("Ours"), and oracle conditions in each of the 12 languages. Steering improves over the unsteered baseline over most languages, with a mean gain of 5.29 ± 1.62 percentage points averaged across languages. Treating each language's seed-averaged improvement as a separate observation, a one-sample t-test against zero improvement gives t = 11.3, p < 0.001, corroborated by a non-parametric Wilcoxon signed-rank test (p < 0.001), a strong, consistent effect by both a parametric and non-parametric measure. 

\begin{table}[t]
\centering
\small
\setlength{\tabcolsep}{4pt}
\begin{tabular}{lccc}
\toprule
Lang. & Baseline & Ours & Oracle \\
\midrule
bn & $56.35\pm2.48$ & $62.70\pm1.37$ & $58.33\pm3.57$ \\
ca & $61.90\pm2.06$ & $69.84\pm1.37$ & $75.40\pm1.37$ \\
de & $69.05\pm3.15$ & $74.21\pm1.37$ & $73.81\pm2.38$ \\
es & $73.02\pm3.83$ & $76.98\pm1.37$ & $76.59\pm1.82$ \\
eu & $31.35\pm3.64$ & $37.30\pm3.00$ & $51.98\pm0.69$ \\
fr & $66.27\pm4.18$ & $71.03\pm2.48$ & $69.44\pm4.18$ \\
gl & $65.87\pm2.75$ & $71.03\pm3.83$ & $76.19\pm1.19$ \\
ja & $55.95\pm3.57$ & $57.94\pm2.48$ & $60.71\pm4.12$ \\
ru & $70.24\pm2.06$ & $73.81\pm2.06$ & $77.78\pm1.37$ \\
sw & $58.33\pm2.38$ & $65.08\pm1.82$ & $66.27\pm0.69$ \\
th & $56.75\pm0.69$ & $63.49\pm2.75$ & $63.89\pm5.63$ \\
zh & $67.06\pm0.69$ & $72.22\pm2.48$ & $75.40\pm3.83$ \\
\bottomrule
\end{tabular}
\caption{Performance across languages over multiple random seeds for the MGSM dataset on the Llama-3.1-8b-Instruct model. Values are mean $\pm$ std.}
\label{tab:seed_variance}
\end{table}

\section{Dendrogram with Task Vectors}
\label{app:task_dendogram}
\begin{figure*}
    \centering
    \includegraphics[width=\linewidth]{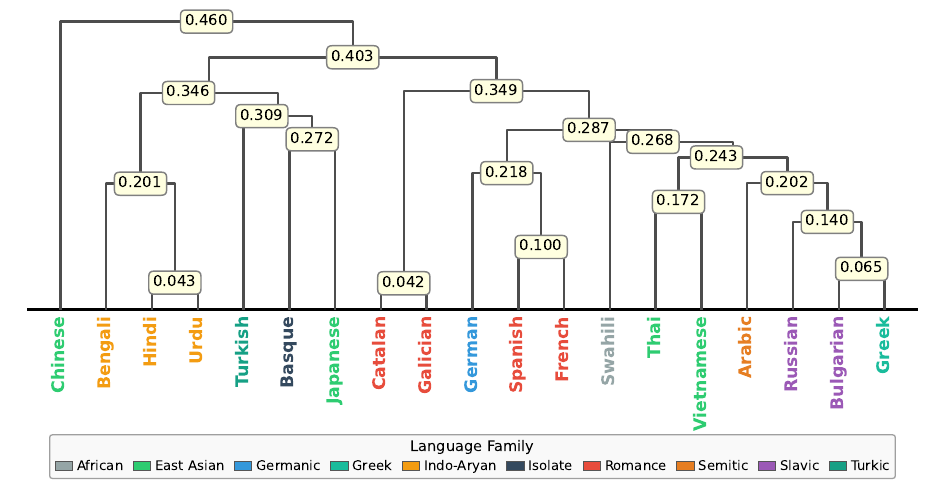}
    \caption{Hierarchical clustering of language vectors computed from task data at layer 10 of \llamamodel. Each language's vector is computed from its available dataset: MGSM for Basque, Catalan, and Galician; XNLI for Arabic, Bulgarian, Greek, Hindi, Turkish, Urdu, and Vietnamese; Edge labels show cosine distances (lower = more similar); colours indicate language families. The closest pairs (Catalan–Galician, Hindi–Urdu, Bulgarian–Greek) match those in the FLORES-based clustering (Figure~\ref{fig:dendogram}), while coarser structure partly reflects the source corpus.}
    \label{fig:dendogram_task}
\end{figure*}

Figure \ref{fig:dendogram_task}, shows the dendrogram calculated over task vectors from different datasets.

\section{Cross-Language Transfer}
Table \ref{tab:crosslingual_near_far_random} tests whether the gains depend on the language direction by replacing each language's vector with vectors from other languages or with norm-matched random vectors, all selected with the same validation protocol. The own-language vector is clearly the most effective, reaching 65.87 on average, compared with 63.59 for the nearest-language vector, 62.43 for random vectors, and 62.10 for the farthest-language vector. It outperforms random vectors in 10 of 12 languages and is the best or tied-best of all 12 language vectors in 8 of 12 languages. The correct language direction thus accounts for most of the improvement (3.4 of 4.9 points), with random perturbation contributing a smaller share (1.4 points). Performance also follows linguistic proximity: vectors from closely related languages recover more of the gain than distant ones, and Catalan's vector nearly matches Galician's own (71.43 vs. 73.81).

\begin{table}[t]
\centering
\small
\begin{tabular}{lrrr}
\toprule
\textbf{Lang} & \textbf{Baseline PPL} & \textbf{Steered PPL} & \textbf{Ratio} \\
\midrule
bn & 1.123 & 1.126 & 1.002 \\
ca & 1.260 & 1.260 & 1.000 \\
de & 1.238 & 1.250 & 1.009 \\
es & 1.231 & 1.233 & 1.002 \\
eu & 1.320 & 1.362 & 1.031 \\
fr & 1.235 & 1.273 & 1.031 \\
gl & 1.258 & 1.280 & 1.017 \\
ja & 1.288 & 1.280 & 0.994 \\
ru & 1.236 & 1.234 & 0.998 \\
sw & 1.248 & 1.269 & 1.017 \\
th & 1.226 & 1.232 & 1.005 \\
zh & 1.268 & 1.272 & 1.003 \\
\midrule
\textbf{Mean} & \textbf{1.244} & \textbf{1.256} & \textbf{1.009} \\
\bottomrule
\end{tabular}
\caption{Perplexity of steered vs. baseline generations, scored under an unsteered forward pass (teacher-forced NLL, response tokens only). A ratio near 1.0 indicates steering does not push generations off the model's natural distribution. Computed for MGSM data on \llamamodel.}
\label{tab:perplexity_sideeffect}
\end{table}

\section{Perplexity Side-Effect Study}
\label{app:perplexity_sideeffect}

To assess whether steering pushes generation away from the model's own distribution, we generate baseline and steered responses on $\mathcal{D}_{\text{test}}$ (greedy decoding, per-language configurations from Table~\ref{tab:best_performing_layers}) and score each response with an unsteered forward pass over the same prompt. Perplexity is computed as $\exp(\overline{\text{NLL}})$ over response tokens only, with the few-shot context and question masked from the loss. Because greedy responses consist of high-probability tokens, absolute perplexities are close to 1; we therefore report the ratio of steered to baseline perplexity per language.

The ratio averages $1.009$ across the 12 languages (range $0.994$--$1.031$; Table~\ref{tab:perplexity_sideeffect}). Steered generations thus remain nearly as probable under the unsteered model as baseline generations, while task accuracy improves by 4.86 points on average. Steering changes which answers the model reaches without moving its outputs off the model's natural distribution. This analysis covers task generations on MGSM.

\section{Prompt Templates}

We describe the exact prompt structure used for each dataset.

\paragraph{MGSM.}
The system message and few-shot demonstrations are passed as the
\textsc{user} turn; the model's generation forms the \textsc{assistant} turn.
The system message is fixed across all languages and experiments:

\begin{quote}
\ttfamily\small
You are a helpful assistant that solves math word problems step by
step. Show your reasoning clearly and end with `Final answer:
\textlangle number\textrangle'.
\end{quote}

The \textsc{user} turn contains $k{=}6$ demonstrations concatenated
with double newlines, each formatted as:

\begin{quote}
\ttfamily\small
Question: \textlangle question in source/target language\textrangle\\
Answer: \textlangle chain-of-thought in English\textrangle\\
Final answer: \textlangle numerical answer\textrangle
\end{quote}

The test question is appended after the few-shot block as:

\begin{quote}
\ttfamily\small
Question: \textlangle question in target language\textrangle\\
Answer:
\end{quote}

\noindent
A key design choice: demonstration \emph{questions} are in the
source language (English) for the baseline, and in the target language
for the oracle. Chain-of-thought reasoning and final answers are
always in English, since the numerical answer is language-invariant.
During steering, demonstration questions remain in English (source
baseline setup) and only the hidden states are modified.

\paragraph{XNLI.}
The system message instructs the model to perform three-way natural
language inference:

\begin{quote}
\ttfamily\small
You are a helpful assistant that performs natural language inference.
Given a premise and a hypothesis, determine the relationship between
them. The relationship can be: `entailment' (hypothesis is true given
the premise), `neutral' (hypothesis might be true or false), or
`contradiction' (hypothesis is false given the premise). Answer with
only: entailment, neutral, or contradiction.
\end{quote}

Each of the $k{=}6$ demonstrations is formatted as:

\begin{quote}
\ttfamily\small
Premise: \textlangle premise\textrangle\\
Hypothesis: \textlangle hypothesis\textrangle\\
Label: \textlangle entailment $|$ neutral $|$ contradiction\textrangle
\end{quote}

The test example is appended as:

\begin{quote}
\ttfamily\small
Premise: \textlangle premise in target language\textrangle\\
Hypothesis: \textlangle hypothesis in target language\textrangle\\
Label:
\end{quote}

\noindent Answer extraction takes only the \emph{first word} of the
model's response, checking for an exact match against the three label
strings.

\paragraph{MSVAMP.}
The prompt structure mirrors MGSM with $k{=}6$ demonstrations.
The system message is identical to MGSM. Each demonstration follows
the same \texttt{Question / Answer / Final answer} format, with
chain-of-thought in English and the numerical answer language-invariant.

%% file: cross_task_new.tex
\begin{table*}[t]
\centering
\setlength{\tabcolsep}{6pt}
\small
\begin{tabular}{lccccclccccc}
\toprule
\multirow{2}{*}{Language} & \multicolumn{5}{c}{MGSM (vector) $\rightarrow$ XNLI (eval)} & & \multicolumn{5}{c}{XNLI (vector) $\rightarrow$ MGSM (eval)} \\
\cmidrule(lr){2-6} \cmidrule(lr){8-12}
 & B & MFS & Ours & CT & \cellcolor{LightGray!30}OR & & B & MFS & Ours & CT & \cellcolor{LightGray!30}OR \\
\midrule
Chinese & 78.14 & 77.84 & 77.84 & \bf{78.14} & \cellcolor{LightGray!30}79.34 & & 84.52 & 88.10 & 86.90 & \bf{90.48} & \cellcolor{LightGray!30}79.76 \\
Thai & 70.96 & \bf{73.05} & 70.06 & 70.06 & \cellcolor{LightGray!30}72.46 & & 79.76 & \bf{82.14} & 79.76 & \bf{82.14} & \cellcolor{LightGray!30}85.71 \\
Swahili & 53.29 & 47.31 & \bf{55.09} & 54.79 & \cellcolor{LightGray!30}54.79 & & 19.05 & 16.67 & \bf{26.19} & 25.00 & \cellcolor{LightGray!30}33.33 \\
Russian & 78.14 & 79.04 & \bf{79.64} & \bf{79.94} & \cellcolor{LightGray!30}78.74 & & 85.71 & 85.71 & \bf{88.10} & \bf{88.10} & \cellcolor{LightGray!30}85.71 \\
French & 81.44 & 82.34 & 82.34 & \bf{82.63} & \cellcolor{LightGray!30}82.63 & & 78.57 & \bf{80.95} & {79.76} & {79.76} & \cellcolor{LightGray!30}84.52 \\
Spanish & 82.63 & 82.63 & \bf{83.53} & 83.23 & \cellcolor{LightGray!30}81.74 & & 84.52 & 83.33 & \bf{89.29} & 88.10 & \cellcolor{LightGray!30}86.90 \\
German & 81.74 & 77.54 & 82.93 & \bf{82.93} & \cellcolor{LightGray!30}79.34 & & 83.33 & 83.33 & \bf{85.71} & \bf{85.71} & \cellcolor{LightGray!30}86.90 \\
\midrule
Average & 75.19 & 74.22 & 75.92 & \bf{75.96} & \cellcolor{LightGray!30}75.58 & & 73.64 & 74.32 & 76.53 & \bf{77.04} & \cellcolor{LightGray!30}77.55 \\
\bottomrule
\end{tabular}

\begin{tabular}{lccccclccccc}
\toprule
\multirow{2}{*}{Language}
& \multicolumn{5}{c}{MGSM (vector) $\rightarrow$ MSVAMP (eval)}
&
& \multicolumn{5}{c}{MSVAMP (vector) $\rightarrow$ MGSM (eval)} \\
\cmidrule(lr){2-6} \cmidrule(lr){8-12}
& B & MFS & Ours & CT & \cellcolor{LightGray!30}OR
&
& B & MFS & Ours & CT & \cellcolor{LightGray!30}OR \\
\midrule
Bengali
& 65.57 & 70.96 & 67.07 & \bf{67.66} & \cellcolor{LightGray!30}76.35
&& 58.33 & \bf{66.67} & 58.33 & 58.33 & \cellcolor{LightGray!30}65.48 \\

German
& 79.04 & 79.94 & \bf{80.24} & 79.64 & \cellcolor{LightGray!30}81.14
&& 83.33 & 83.33 & \bf{85.71} & 77.38 & \cellcolor{LightGray!30}86.90 \\

Spanish
& 82.04 & 83.53 & \bf{83.83} & 82.04 & \cellcolor{LightGray!30}84.13
&& 84.52 & 83.33 & 89.29 & \bf{90.48} & \cellcolor{LightGray!30}86.90 \\

French
& 82.34 & \bf{85.03} & 84.13 & 79.04 & \cellcolor{LightGray!30}84.43
&& 78.57 & \bf{80.95} & 79.76 & 77.38 & \cellcolor{LightGray!30}84.52 \\

Japanese
& 78.74 & \bf{79.64} & \bf{79.64} & 79.34 & \cellcolor{LightGray!30}79.94
&& 71.43 & 75.00 & 75.00 & \bf{76.19} & \cellcolor{LightGray!30}73.81 \\

Russian
& 79.64 & \bf{82.93} & 81.74 & 80.54 & \cellcolor{LightGray!30}81.44
&& 85.71 & 85.71 & \bf{88.10} & 83.33 & \cellcolor{LightGray!30}85.71 \\

Swahili
& 49.10 & 47.90 & \bf{50.60} & 48.50 & \cellcolor{LightGray!30}50.30
&& 19.05 & 16.67 & \bf{26.19} & 20.24 & \cellcolor{LightGray!30}33.33 \\

Thai
& 78.14 & \bf{79.34} & \bf{79.34} & 76.65 & \cellcolor{LightGray!30}78.74
&& 79.76 & \bf{82.14} & 79.76 & 78.57 & \cellcolor{LightGray!30}85.71 \\

Chinese
& 82.04 & 83.23 & \bf{83.53} & 83.23 & \cellcolor{LightGray!30}82.34
&& 84.52 & \bf{88.10} & 86.90 & \bf{88.10} & \cellcolor{LightGray!30}79.76 \\

\midrule
Average
& 75.18 & \bf{76.94} & 76.68 & 75.18 & \cellcolor{LightGray!30}77.65
&& 71.69 & 73.54 & \bf{74.34} & 72.22 & \cellcolor{LightGray!30}75.79 \\
\bottomrule
\end{tabular}

\begin{tabular}{lccccclccccc}
\toprule
\multirow{2}{*}{Language}
& \multicolumn{5}{c}{MSVAMP (vector) $\rightarrow$ XNLI (eval)}
&
& \multicolumn{5}{c}{XNLI (vector) $\rightarrow$ MSVAMP (eval)} \\
\cmidrule(lr){2-6} \cmidrule(lr){8-12}
& B & MFS & Ours & CT & \cellcolor{LightGray!30}OR
&
& B & MFS & Ours & CT & \cellcolor{LightGray!30}OR \\
\midrule
Chinese
& 78.14 & 77.84 & 77.84 & \bf{78.74} & \cellcolor{LightGray!30}79.34
&& 82.04 & 83.23 & \bf{83.53} & 80.84 & \cellcolor{LightGray!30}82.34 \\

Thai
& 70.96 & \bf{73.05} & 70.06 & 67.96 & \cellcolor{LightGray!30}72.46
&& 78.14 & \bf{79.34} & \bf{79.34} & 76.35 & \cellcolor{LightGray!30}78.74 \\

Swahili
& 53.29 & 47.31 & \bf{55.09} & 52.99 & \cellcolor{LightGray!30}54.79
&& 49.10 & 47.90 & \bf{50.60} & 48.80 & \cellcolor{LightGray!30}50.30 \\

Russian
& 78.14 & 79.04 & \bf{79.64} & 76.05 & \cellcolor{LightGray!30}78.74
&& 79.64 & \bf{82.93} & 81.74 & 79.64 & \cellcolor{LightGray!30}81.44 \\

French
& 81.44 & \bf{82.34} & \bf{82.34} & 80.24 & \cellcolor{LightGray!30}82.63
&& 82.34 & \bf{85.03} & 84.13 & 81.44 & \cellcolor{LightGray!30}84.43 \\

Spanish
& 82.63 & 82.63 & \bf{83.53} & 82.04 & \cellcolor{LightGray!30}81.74
&& 82.04 & 83.53 & \bf{83.83} & 82.63 & \cellcolor{LightGray!30}84.13 \\

German
& 81.74 & 77.54 & \bf{82.93} & 80.24 & \cellcolor{LightGray!30}79.34
&& 79.04 & 79.94 & 80.24 & \bf{81.14} & \cellcolor{LightGray!30}81.14 \\

\midrule
Average
& 75.19 & 74.25 & \bf{75.92} & 74.04 & \cellcolor{LightGray!30}75.58
&& 76.05 & 77.41 & \bf{77.63} & 75.83 & \cellcolor{LightGray!30}77.50 \\
\bottomrule

\end{tabular}

\caption{Cross-task transfer results across all task-vector combinations between MGSM, MSVAMP, and XNLI on \qwenseven. For each task pair, we compute steering vectors from one task and evaluate on another task. B = Baseline, MFS = Multilingual Few-Shot Baseline \cite{tu2025blessing}, CT = Cross-Transfer, OR = Oracle (target-language few-shot reference, not intended for direct comparison). \textbf{Bold} indicates the best score.}
\label{tab:cross_task_corrected}
\end{table*}